\documentclass{article}
\pdfoutput=1

\usepackage{selectp}

\usepackage{microtype}
\usepackage{graphicx}
\usepackage{booktabs} 

\usepackage{hyperref}

\usepackage[accepted]{icml2021}

\usepackage{url}

\usepackage[utf8]{inputenc} 
\usepackage[T1]{fontenc}    
\usepackage{hyperref}       
\usepackage{url}            
\usepackage{amsfonts}       
\usepackage{amsbsy}
\usepackage{nicefrac}       
\usepackage{authblk}
\usepackage{subcaption}
\usepackage{datetime}
\usepackage{multirow}
\usepackage{xcolor}
\usepackage{amsmath}
\usepackage{float}
\usepackage[bottom]{footmisc}
\usepackage{algorithm}
\usepackage[noend]{algorithmic}
\usepackage{natbib}
\usepackage{graphics}
\usepackage{multirow}
\usepackage{mathtools}
\usepackage{nicefrac}

\usepackage{wrapfig}
\usepackage[export]{adjustbox}
\usepackage{bbm}
\usepackage{comment}

\usepackage{dblfloatfix}

\newcommand*\diff{\mathop{}\!\mathrm{d}}

\DeclareMathOperator*{\argmax}{arg\,max}

\newcommand{\met}{HO2}
\newcommand{\method}{Hindsight Off-policy Options}

\icmltitlerunning{Data-efficient Hindsight Off-policy Option Learning}

\begin{document}

\twocolumn[
\icmltitle{Data-efficient Hindsight Off-policy Option Learning}



\icmlsetsymbol{equal}{*}

\begin{icmlauthorlist}
\icmlauthor{Markus Wulfmeier}{dm}
\icmlauthor{Dushyant Rao}{dm}
\icmlauthor{Roland Hafner}{dm}
\icmlauthor{Thomas Lampe}{dm}
\icmlauthor{Abbas Abdolmaleki}{dm}
\icmlauthor{Tim Hertweck}{dm}
\icmlauthor{Michael Neunert}{dm}
\icmlauthor{Dhruva Tirumala}{dm}
\icmlauthor{Noah Siegel}{dm}
\icmlauthor{Nicolas Heess}{dm}
\icmlauthor{Martin Riedmiller}{dm}
\end{icmlauthorlist}

\icmlaffiliation{dm}{DeepMind, London, United Kingdom}

\icmlcorrespondingauthor{Markus Wulfmeier}{mwulfmeier@google.com}

\icmlkeywords{Hierarchical Reinforcement Learning, Off-policy Learning, Abstraction, Probabilistic Inference}

\vskip 0.3in
]



\printAffiliationsAndNotice{}  

\begin{abstract}
We introduce \method~(\met), a data-efficient option learning algorithm.
Given any trajectory, \met~infers likely option choices and backpropagates through the dynamic programming inference procedure to robustly train all policy components off-policy and end-to-end.
The approach outperforms existing option learning methods on common benchmarks. 
To better understand the option framework and disentangle benefits from both temporal and action abstraction, we evaluate ablations with flat policies and mixture policies with comparable optimization.
The results highlight the importance of both types of abstraction as well as off-policy training and trust-region constraints, particularly in challenging, simulated 3D robot manipulation tasks from raw pixel inputs.
Finally, we intuitively adapt the inference step to investigate the effect of increased temporal abstraction on training with pre-trained options and from scratch.
\end{abstract}

\section{Introduction}\label{sec:introduction}

Deep reinforcement learning has seen numerous successes in recent years~\citep{silver2017mastering,openai2018dexterous,vinyals2019grandmaster}, but still faces challenges in domains where data is limited or expensive. 
One candidate solution to address the challenges and improve data efficiency is to impose hierarchical policy structures.
By dividing an agent into a combination of low-level and high-level controllers, the options framework \citep{sutton1999between,precup2000temporal} introduces a form of action abstraction, effectively reducing the high-level controller's task to choosing from a discrete set of reusable sub-policies.
The framework further enables temporal abstraction by explicitly modelling the temporal continuation of low-level behaviors.
Unfortunately, in practice, hierarchical control schemes often introduce technical challenges, including a tendency to learn degenerate solutions preventing the agent from using its full capacity \citep{ harb2018waiting}, undesirable trade-offs between learning efficiency and final performance \citep{harut2019termination}, or the increased variance of updates  \citep{precup2000temporal}.
Additional challenges in off-policy learning for hierarchical approaches \citep{NIPS2005_2767} led to a focus of recent works on the on-policy setting, forgoing the considerable improvements in data efficiency often connected to off-policy methods.

\begin{figure*}[t]
	\centering
	\begin{tabular}{ccc}
	    \includegraphics[trim=0 3mm 0 0,clip,width = 0.29\textwidth,valign=b]{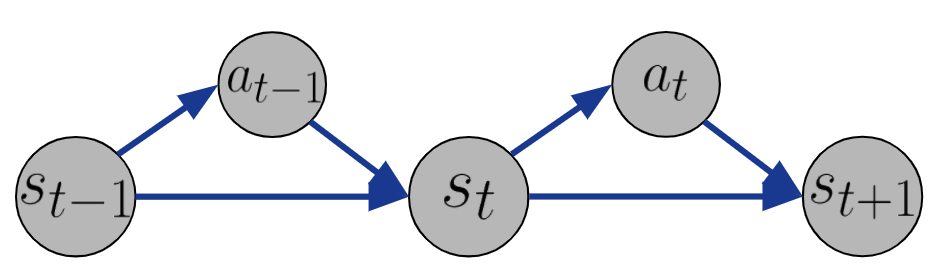} &
        \includegraphics[trim=0 9mm 0 0,clip,width = 0.32\textwidth,valign=b]{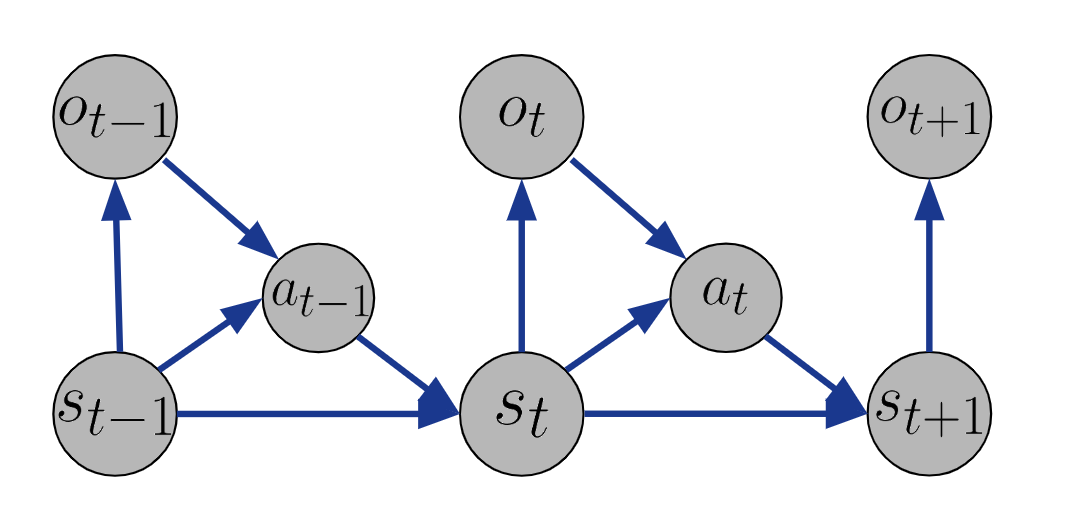} &
        \includegraphics[width = 0.29\textwidth,valign=b]{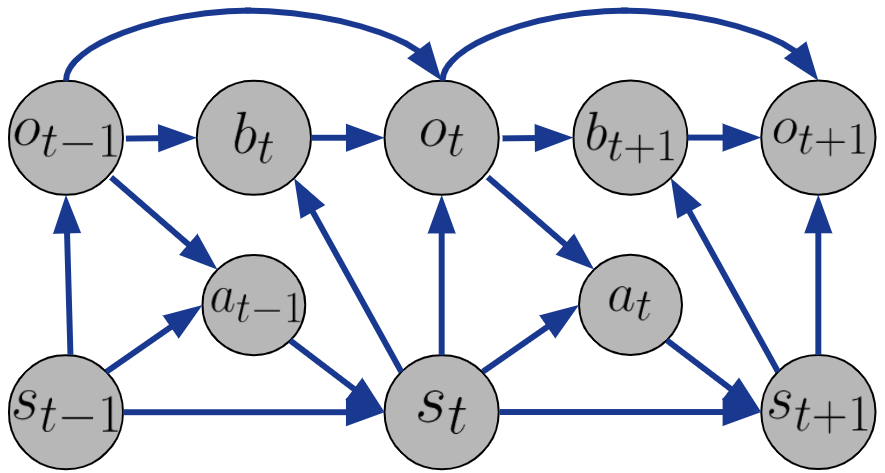}
	\end{tabular}
	\vspace{-1mm}
    \caption{Graphical model for flat policies (left), mixture policies (middle) - introducing a type of action abstraction, and option policies (right) - adding temporal abstraction via autoregressive options.
    While the action $a$ is solely dependent on the state $s$ for flat policies, mixture policies introduce the additional component or option $o$ which affects the actions (following Equation \ref{eq:mixture}). Option policies do not change the direct dependencies for actions but instead affect the options themselves, which are now also dependent on the previous option and its potential termination $b$ (following Equation \ref{eq:optiontransitions}). } 
	\label{fig:2dimensional}
	\vspace{-1mm}
\end{figure*}

We propose an approach to address these drawbacks, \method~(\met): a method for data-efficient, robust, off-policy option learning.
The algorithm simultaneously learns a high-level controller and low-level options via a single end-to-end optimization procedure. It improves data efficiency by leveraging off-policy learning and inferring distributions over option for trajectories in hindsight to maximize the likelihood of good actions and options.

To facilitate off-policy learning the algorithm does not condition on executed options but treats these as latent variables during optimization and marginalizes over all options to compute the exact likelihood.
\met~backpropagates through the resulting dynamic programming inference graph (conceptually related to \citep{rabiner1989tutorial,shiarlis2018taco, smith2018inference}) to enable the training of all policy components from trajectories, independent of the executed option.
As an additional benefit, the formulation of the inference graph allows to impose intuitive, hard constraints on the option termination frequency, thereby regularizing the learned solution (and encouraging temporally-extended behaviors) independently of the scale of the reward.

The policy update follows an expectation-maximization perspective and generates an intermediate, non-parametric policy, which is adapted to maximize agent performance. This enables the update of the parametric policy to rely on simple weighted maximum likelihood, without requiring further approximations such as Monte Carlo estimation or continuous relaxation~\citep{li2019sub}. 
Finally, the updates are stabilized using adaptive trust-region constraints, demonstrating the importance of robust policy optimization for hierarchical reinforcement learning (HRL) in line with recent work on on-policy option learning \citep{zhang2019dac}.

We experimentally compare \met~to prior option learning methods.
By treating options as latent variables in off-policy learning and enabling backpropagation through the inference procedure, \met~demonstrates to be more efficient than prior approaches such as the Option-Critic \citep{bacon2017option} or DAC \citep{zhang2019dac}. \met additionally outperforms IOPG \citep{smith2018inference}, which considers a similar perspective but still builds on on-policy training.
To better understand different abstractions in option learning, we compare with corresponding policy optimization methods for flat policies \citep{abdolmaleki2018relative} and mixture policies without temporal abstraction \citep{wulfmeier2019regularized}
thereby allowing us to isolate the benefits of both action and temporal abstraction.
Both properties demonstrate particular relevance in more demanding simulated robot manipulation tasks from raw pixel inputs.
We further perform extensive ablations to evaluate the impact of trust-region constraints, off-policyness, option decomposition, and the benefits of maximizing temporal abstraction when using pre-trained options versus learning from scratch.

Our main contributions include:
\begin{itemize}
    \item A robust, efficient off-policy option learning algorithm enabled by a probabilistic inference perspective on HRL. The method outperforms existing option learning methods on common benchmarks and demonstrates benefits on pixel-based 3D robot manipulation tasks.
    \item An intuitive technique to further encourage temporal abstraction beyond the core method, using the inference graph to constrain option switches without additional weighted loss terms.
    \item A careful analysis to improve our understanding of the options framework by isolating the impact of action abstraction and temporal abstraction.
    \item Further ablation and analysis of several algorithmic choices: trust-region constraints, off-policy versus on-policy data, option decomposition, and the importance of temporal abstraction with pre-trained options versus learning from scratch.
\end{itemize}

\section{Method}\label{sec:method}\label{sec:prelims}

We start by considering a reinforcement learning setting with an agent operating in a Markov Decision Process (MDP) consisting of the state space $\mathcal{S}$, the action space $\mathcal{A}$, and the transition probability $p(s_{t+1}|s_t,a_t)$ of reaching state $s_{t+1}$ from state $s_t$ when executing action $a_t$. 
The agent's behavior is commonly described as a conditional distribution with actions $a_t$ drawn from the agent's policy $\pi(a_t | s_t)$.
Jointly, the transition dynamics and policy induce the marginal state visitation distribution $p(s_t)$. The discount factor $\gamma$ together with the reward $r_t=r\left(s_{t}, a_{t}\right)$ gives rise to the expected return, which the agent aims to maximize:
$J(\pi) = \mathbb{E}_{p\left(s_t\right),\pi(a_t|s_t)}\Big[\sum_{t=0}^{\infty} \gamma^{t} r_t \Big]$.

\subsection{Policy Types} 
Option policies introduce temporal and action abstraction in comparison to commonly-used flat Gaussian policies.
Our goal in this work is not only to introduce this additional structure to improve data efficiency but to further understand the impact of the different abstractions. 
For this purpose, we further study mixture distributions. They represent an intermediate case with only action abstraction, as described in Figure \ref{fig:2dimensional}. 

We begin by covering both policy types in the following paragraphs. First, we focus on computing likelihoods of actions (and options) under a policy. Then, we describe the proposed critic-weighted maximum likelihood algorithm to train hierarchical policies.

\paragraph{Mixture Policies} This type extends flat policies $\pi(a_t | s_t)$ by introducing a high-level controller that samples from multiple options (low-level policies) independently at each timestep (Figure \ref{fig:2dimensional}).
The joint probability of actions and options 
is given as: 
\begin{align}\label{eq:mixture}
\pi_{\theta}(a_t,o_t | s_t) = \pi^L \left(a_t | s_t, o_t\right) \pi^H\left(o_t | s_t\right), 
\end{align}
where $\pi^H$ and $\pi^L$ respectively represent high-level policy (which for the mixture is equal to a Categorical distribution $\pi^H\left(o_t | s_t\right) = \pi^C\left(o_t | s_t\right)$) and low-level policy (components of the resulting mixture distribution), and $o$ is the index of the sub-policy or mixture component. 

\paragraph{Option Policies} This type extends mixture policies by incorporating temporal abstraction.
We follow the semi-MDP and \textit{call-and-return} option model \citep{sutton1999between}, defining an option as a triple $(I(s_t,o_t),\pi^L(a_t|s_t,o_t),\beta(s_t,o_t))$. The initiation condition $I$ describes an option's probability to start in a state and is simplified to $I(s_t,o_t)=1 \forall s_t \in \mathcal{S}$ following \citep{bacon2017option,zhang2019dac}. The termination condition $b_t \sim \beta(s_t, o_t)$ denotes a Bernoulli distribution describing the option's probability to terminate in any given state, and the action distribution for a given option is modelled by $\pi^L(a_t|s_t,o_t)$.
Every time the agent observes a state, the current option's termination condition is sampled. If subsequently no option is active, a new option is sampled from the controller $\pi^C(o_{t}|s_{t})$. Finally, we sample from either the continued or new option to generate a new action. 
The resulting transition probabilities between options are described by 
\begin{align}
p\left(o_t | s_t, o_{t-1}\right) =~~~~~~~~~~~~~~~~~~~~~~~~~~~~~~~~~~~~~~~~~~~~~~~~~~~~~~~\label{eq:optiontransitions}
 \\ 
  \nonumber\begin{cases} 
    1-\beta(s_t,o_{t-1}) (1- \pi^C(o_t|s_t)) & \text{if } o_{t} = o_{t-1} \\
   \beta(s_t,o_{t-1}) \pi^C(o_t|s_t)      & \text{otherwise}
  \end{cases}
\end{align}
During interaction in an environment, an agent samples individual options. However, during learning \met~takes a probabilistic inference perspective with options as latent variables and states and actions as observed variables. 
This allows us to infer likely options over a whole trajectory in hindsight, leading to efficient intra-option learning \citep{precup2000temporal} for all options independently of the executed option. This is particularly relevant for off-policy learning, as options can change between data generation and learning.

Following the graphical model in Figure \ref{fig:2dimensional} and corresponding transition probabilities in Equation \ref{eq:optiontransitions}, the probability of being in option $o_t$ at timestep $t$ across a trajectory $h_t= \{s_t, a_{t-1}, s_{t-1},... s_0, a_0\}$ is determined in a recursive manner based on the previous timestep's option probabilities. For the first timestep, the probabilities are given by the high-level controller $\pi^{H}\left(o_0 | h_0\right) = \pi^C\left(o_0 | s_0\right)$. For all consecutive steps are computed as follows for $M$ options:
\begin{eqnarray}
\begin{aligned}
     \tilde{\pi}^{H}\left(o_t | h_t\right) = \sum_{o_{t-1}=1}^M \big[ p\left(o_t|s_t, o_{t-1}\right)  \pi^H\left(o_{t-1} | h_{t-1}\right) \label{eq:dynamic1} \\
     {\pi^L\left(a_{t-1} | s_{t-1}, o_{t-1}\right)}\big] 
\end{aligned}
\end{eqnarray}
The distribution is normalized at each timestep following $\pi^H\left(o_t | h_t\right) = {\tilde{\pi}^{H}\left(o_t | h_t\right)}/ {\sum_{o'_{t}=1}^M\tilde{\pi}^{H}\left(o'_t | h_t\right)}$.
Performing this exact marginalization at each timestep is much more efficient than computing independently over all possible sequences of options and reduces variance compared to sampling-based approximations.

Building on the option probabilities, Equation \ref{eq:option_probs} conceptualizes the connection between mixture and option policies. 
\begin{align}\label{eq:option_probs}
\pi_{\theta}(a_t,o_t | h_{t}) =  \pi^L\left(a_t | s_t, o_t\right) \pi^H\left(o_t | h_t\right)
\end{align}
In both cases, the low-level policies $\pi^L$ only depend on the current state.
However, where mixtures only depend on the current state $s_t$ for the high-level probabilities $\pi^H$, with options we can take into account compressed information about the history $h_{t}$ as facilitated by the previous timestep's distribution over options $\pi^H\left(o_{t-1} | h_{t-1}\right)$.

This dynamic programming formulation in Equation \ref{eq:dynamic1} 
enables the exact computation of the likelihood of actions and options along off-policy trajectories.
We can use automatic differentiation in modern deep learning frameworks (e.g. \citep{abadi2016tensorflow}) to backpropagate through the graph and determine the gradient updates for all policy parameters.

\subsection{Agent Updates} 
We continue by describing the policy improvement algorithm, which uses the previously determined option probabilities. 
The three main steps are: 1) update the critic (Eq. \ref{eq:objective_q_value}); 2) generate an intermediate, non-parametric policy based on the updated critic (Eq. \ref{eq:objective_q}); 3) update the parametric policy to align to the non-parametric improvement (Eq. \ref{eq:objective_pi}). 
By handling the maximization of expected returns with a closed-loop solution for a non-parametric intermediate policy, the update of the parametric policy can build on simple, weighted maximum likelihood.
In essence, we do not rely on differentiating an expectation over a distribution with respect to parameters of the distribution.
This enables training a broad range of distributions (including discrete ones) without further approximations such as required when the update relies on the reparametrization trick \citep{li2019sub}. 
\begin{figure*}[t]
	\centering
	\begin{tabular}{cc}
        \includegraphics[width = 0.43\textwidth]{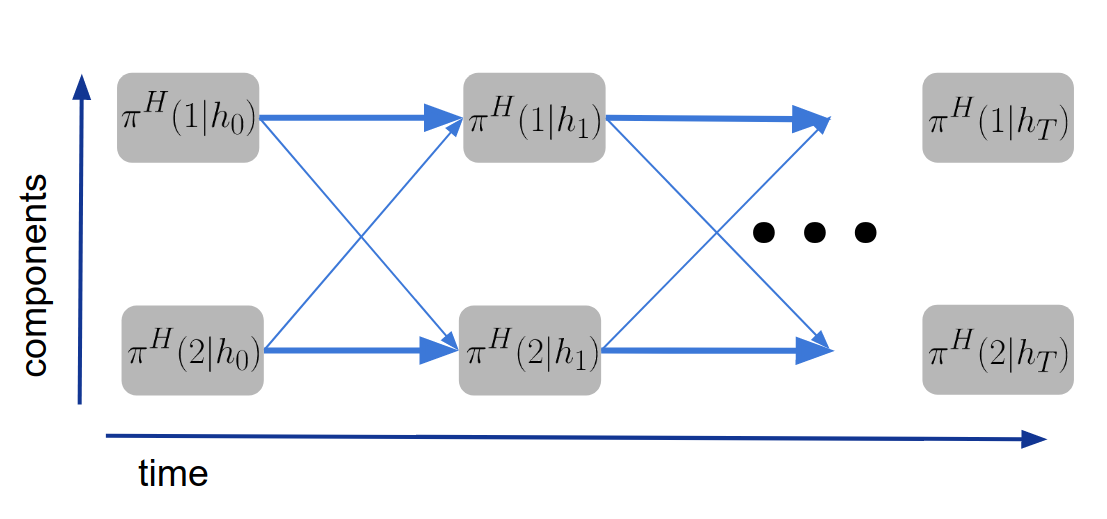}&
        \includegraphics[width = 0.53\textwidth]{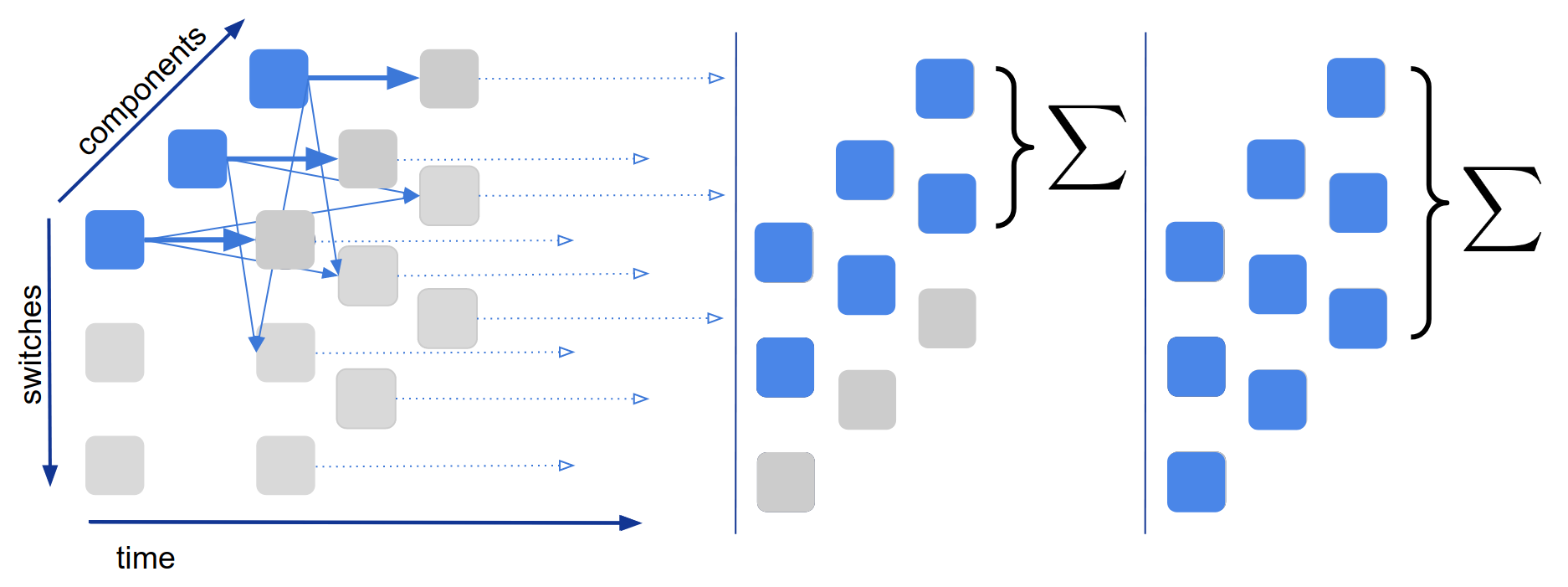}  
	\end{tabular}
    \caption{Representation of the dynamic programming forward pass - bold arrows represent connections without switching. Left: example with two options. Right: extension of the graph to explicitly count the number of switches. Marginalization over the dimension of switches determines component probabilities. By limiting over which nodes to sum at every timestep, the optimization can be targeted to fewer switches and more consistent option execution.}
	\label{fig:example}
\end{figure*}

\paragraph{Policy Evaluation} 
In comparison to prior work on training mixture policies~\citep{wulfmeier2019regularized}, the critic for option policies is a function of $s$, $a$, {and} $o$ since the current option 
influences the likelihood of future actions and thus rewards. Note that even though we express the policy as a function of the history $h_t$, $Q$ is a function of $o_t, s_t, a_t$, since these are sufficient to render the future trajectory independent of the past (see the graphical model in Figure \ref{fig:2dimensional}).
We define the TD(0) objective as 
\begin{align}
\label{eq:objective_q_value}
\min_\phi L(\phi) =  \mathbb{E}_{s_t,a_t,o_t \sim \mathcal{D}} \Big[  \big( Q_{\text{T}} - {Q}_\phi(s_t, a_t, o_t))^2 \Big],
\end{align}
where the current states, actions, and options are sampled from the current replay buffer $\mathcal{D}$. 
For the 1-step target $Q_{\text{T}}= r_t+\gamma \mathbb{E}_{s_{t+1},a_{t+1},o_{t+1}}[{{Q'}}(s_{t+1}, a_{t+1}, o_{t+1})]$, 
the expectation over the next state is approximated with the sample $s_{t+1}$ from the replay buffer,
and we estimate the value by sampling actions and options according to $a_{t+1}, o_{t+1} \sim \pi'(\cdot|h_{t+1})$  following Equation \ref{eq:option_probs}. $\pi'$ and $Q'$ respectively represent target networks for policy and critic which are used to stabilize training.

\paragraph{Policy Improvement}
We follow an Expectation-Maximization procedure similar to \citep{wulfmeier2019regularized,abdolmaleki2018maximum}, which first computes an improved non-parametric policy and then updates the parametric policy to match this target. 
In comparison to prior work, the policy does not only depend on the current state $s_t$ but also on compressed information about the rest of the previous trajectory $h_t$, building on Equation \ref{eq:dynamic1}.
Given the critic, all we require to optimize option policies is the ability to sample from the policy and determine the log-likelihood (gradient) of actions and options under the policy. 
The first step provides us with a non-parametric policy $q(a_t,o_t|h_t)$. 
\begin{eqnarray}
  \begin{aligned}
    \max_{q} J(q) =\ \mathbb{E}_{a_t,o_t \sim q, h_t \sim \mathcal{D}}\big[{Q_\phi}(s_t, a_t, o_t) \big], \\ 
    \text{s.t. } \mathbb{E}_{h_t \sim \mathcal{D}}\Big[
    \mathrm{KL}\big(q(\cdot|h_t) \| \pi_\theta(\cdot|h_t)\big)\Big] \le \epsilon_E,
  \end{aligned}
  \label{eq:objective_q}
\end{eqnarray}
where $\mathrm{KL}(\cdot \| \cdot)$ denotes the Kullback-Leibler divergence, and $\epsilon_E$ defines a bound on the KL.
We can find the solution to the constrained optimization problem in Equation \ref{eq:objective_q} in closed-form and obtain
\begin{equation} \label{eq:q_closed}
q(a_t,o_t|h_t) \propto \pi_{\theta}(a_t,o_t|h_t) \exp\left({{{Q_\phi}(s_t,a_t,o_t)}/{\eta}}\right).
\end{equation}
Practically speaking, this step computes samples from the previous policy and weights these based on the corresponding temperature-calibrated values of the critic. The temperature parameter $\eta$ is computed following the dual of the Lagrangian. The derivation and final form of the dual can be found in Appendix \ref{app:nonparam}, Equation \ref{eq:dual_eta}. 

To align the parametric to the improved non-parametric policy in the second step, we minimize their KL divergence. 
\begin{eqnarray}
\begin{aligned}
    \theta = \arg \min_{\theta} \mathbb{E}_{h_t \sim \mathcal{D}}\Big[ \mathrm{KL}\big( q(\cdot | h_t) \| \pi_{\theta}(\cdot | h_t) \big) \Big], \label{eq:objective_pi}\\
    \text{s.t. } \mathbb{E}_{h_t \sim \mathcal{D}}\Big[
    \mathcal{T}\big(\pi_{\theta_{+}}(\cdot|h_t) \| \pi_{\theta}(\cdot|h_t)\big)\Big] \le \epsilon_M
\end{aligned}
\end{eqnarray}
The distance function $\mathcal{T}$ in Equation \ref{eq:objective_pi} has a trust-region effect and stabilizes learning by constraining the change in the parametric policy. The computed option probabilities from Equation \ref{eq:dynamic1} are used in Equation \ref{eq:q_closed} to enable sampling of options as well as Equation \ref{eq:objective_pi} to determine and maximize the likelihood of samples under the policy.
We can apply Lagrangian relaxation again and solve the primal as detailed in Appendix \ref{app:param}.
Finally, we describe the complete pseudo-code for \met~in Algorithm \ref{alg:learner}.

Note that both Gaussian and mixture policies have been trained in prior work via methods relying on critic-weighted maximum likelihood \citep{abdolmaleki2018relative, wulfmeier2019regularized}. By comparing with the extension towards option policies described above, we will make use of this connection to isolate the impact of action abstraction and temporal abstraction in the option framework in Section \ref{sec:multi}.

\begin{algorithm}[t]
\caption{\method}\label{alg:learner}
\begin{algorithmic}
\STATE \textbf{input:} initial parameters for $\theta,~\eta$ and $\phi$, KL regularization parameters $\epsilon$, set of trajectories $\tau$
\WHILE{not done}
\STATE sample trajectories $\tau$ from replay buffer
\STATE \color{gray}// forward pass along sampled trajectories
\STATE \color{black}determine component probabilities $\pi^H\left(o_t | h_t\right)$ (Eq. \ref{eq:dynamic1})
\STATE sample actions $a_j$ and options $o_j$ from $\pi_{\theta}(\cdot|h_t)$ (Eq. \ref{eq:option_probs}) to estimate expectations
\STATE \color{gray}// compute gradients over batch for policy, Lagrangian multipliers and Q-function
\STATE \color{black}$\delta_\theta \leftarrow -\nabla_\theta \sum_{h_t \in \tau} \sum_{j} [ \exp\left({Q_\phi(s_t,a_j,o_j)}/{\eta}\right)$ \\ \hfill $\log \pi_{\theta}(a_j, o_j | h_t) ] $ following Eq. \ref{eq:q_closed} and \ref{eq:objective_pi}
\STATE $\delta_{\eta} \leftarrow \nabla_\eta g(\eta) = \nabla_\eta \eta\epsilon+\eta \sum_{h_t \in \tau} \log  \sum_{j}[$ \\ \hfill $\exp\left({Q_\phi(s_t,a_j,o_j)}/{\eta}\right) ]$ following Eq. \ref{eq:dual_eta}
\STATE $\delta_\phi \leftarrow \nabla_{\phi} \sum_{(s_t, a_t, o_t) \in \tau} \big( {Q}_\phi(s_t, a_t, o_t) -
  Q_{\text{T}} \big)^2$ \\ \hfill following Eq. \ref{eq:objective_q_value}
\STATE \color{black}update $\theta, \eta, \phi$  \color{gray}~~~~~~// apply gradient updates \color{black}
\IF{number of iterations = target update}
\STATE \color{black}$\pi' = \pi_\theta$, $Q' = Q_\phi$ \color{gray}~~~~~~// update target networks for policy $\pi'$ and value function $Q'$   \color{black}
\ENDIF
\ENDWHILE
\end{algorithmic}
\end{algorithm}

\begin{figure*}[b]
	\centering
	\begin{tabular}{c}
        \includegraphics[trim=3mm 0 0 0, clip,width = 0.98\textwidth]{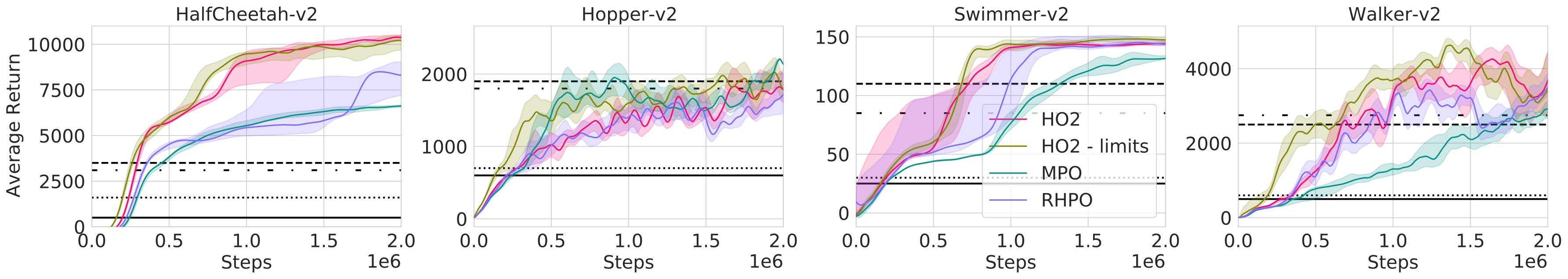} 
	\end{tabular}
    \caption{Results on OpenAI gym. Dashed black line represents DAC \citep{zhang2019dac}, dotted line represents Option-Critic \citep{bacon2017option}, solid line represents IOPG \citep{smith2018inference}, dash-dotted line represents PPO \citep{schulman2017proximal} (approximate results after $2 \times 10^{6}$ steps from \citep{zhang2019dac}). We limit the number of switches to 5 for \met-limits. \met~ consistently performs better or on par with existing option learning algorithms. }
	\label{fig:openaigym}
\end{figure*}

\subsection{Maximizing Temporal Abstraction}
\label{sec:temporal_abstraction}
Persisting with each option over longer time periods can help to reduce the search space and simplify exploration \citep{sutton1999between,harb2018waiting}.
Previous approaches (e.g. \citep{harb2018waiting}) rely on additional weighted loss terms which penalize option transitions.

In addition to the main \met~algorithm, we introduce an extension mechanism to explicitly limit the maximum number of switches between options along a trajectory to increase temporal abstraction.
In comparison to additional loss terms, a parameter for the maximum number of switches can be chosen independently of the reward scale of an environment and provides an intuitive semantic interpretation, both aspects simplify manual adaptation.

We extend the 2D graph for computing option probabilities (Figure \ref{fig:example}) with a third dimension representing the number of switches between options. Practically, this means that we are modelling $\pi^H(o_t,n_t|h_t)$ where $n_t$ represents the number of switches until timestep $t$. 
We can now marginalize over \textit{all} numbers of switches to again determine the option probabilities. Instead, to encourage option consistency across timesteps, we can sum over \textit{only a subset} of nodes for all $n \leq  N$ with $N$ smaller than the maximal number of switches leading to $\pi^H \left(o_t| h_t\right) = \sum_{n_t=0}^N \pi^H  \left(o_t, n_t| h_t\right)$.

For the first timestep, only 0 switches are possible, such that $\pi^H\left(o_0 ,n_0=0| h_0\right) = \pi^C\left(o_0 |  s_0\right)$ and $0$ for all other values of $n$.
For further timesteps, all edges resulting in option terminations $\beta$ lead to the next step's option probabilities with increased number of switches $n_{t+1} = n_t +1$. All edges representing the continuation of an option lead to $n_{t+1} = n_t $.
This results in the computation of the joint distribution for $t>0$:
\begin{eqnarray}
\begin{aligned}
    \tilde{\pi}^{H}\left(o_t,n_t | h_t\right) = \sum_{\substack{o_{t-1}=1,\\ n_{t-1}=1}}^{M,N}  p\left(o_t,n_t|s_t, o_{t-1},n_{t-1}\right) \\ \pi^H\left(o_{t-1},n_{t-1} | h_{t-1}\right) \label{eq:dynamic2}
    {\pi^L\left(a_{t-1} | s_{t-1}, o_{t-1}\right)} %
\end{aligned}
\end{eqnarray}
\noindent which can then be normalized using $\pi^H\left(o_t,n_t | h_t\right) = {\tilde{\pi}^{H}\left(o_t,n_t | h_t\right)}/ {\sum_{o'_{t}=1}^M\sum_{n'_{t}=1}^L\tilde{\pi}^{H}\left(o'_t ,n'_t| h_t\right)}$.
The option and switch index transitions $p\left(o_t,n_t|s_t, o_{t-1},n_{t-1}\right)$ are further described in Equation \ref{eq:optionswitch_transitions} in the Appendix.
\section{Experiments}\label{sec:experiments}

In this section, we aim to answer a set of questions to better understand the contribution of different aspects to the performance of option learning - in particular with respect to the proposed method, \met. 
To start, in Section \ref{sec:single} we explore two questions: (1) How well does HO2 perform in comparison to existing option learning methods? and (2) How important is off-policy training in this context?   
We use a set of common OpenAI gym \citep{Brockman2016OpenAIG} benchmarks to answer these questions.
In Section \ref{sec:multi} we ask: (3) How do action abstraction in mixture policies and the additional temporal abstraction brought by option policies individually impact performance?
We use more complex, pixel-based 3D robotic manipulation experiments to investigate these two aspects and evaluate scalability with respect to higher dimensional input and state spaces.
We also explore: (4) How does increased temporal consistency impact performance, particularly with respect to sequential transfer with pre-trained options? 
Finally, we perform additional ablations in Section \ref{sec:ablations} to investigate the challenges of robust off-policy option learning and improve understanding of how options decompose behavior based on various environment and algorithmic aspects.

Across domains, we use \met~to train option policies, RHPO \citep{wulfmeier2019regularized} for the reduced case of mixture-of-Gaussians policies with sampling of options at every timestep and MPO \citep{abdolmaleki2018relative} to train individual flat Gaussian policies - all based on critic-weighted maximum likelihood estimation for policy optimization.

\subsection{Comparison of Option Learning Methods} \label{sec:single}
We compare \met~(with and without limits on option switches) against competitive baselines for option learning in common, feature-based continuous action space domains. \met~outperforms baselines including Double Actor-Critic (DAC) \citep{zhang2019dac}, Inferred Option Policy Gradient (IOPG) \citep{smith2018inference} and Option-Critic (OC) \citep{bacon2017option}.
With PPO~\citep{schulman2017proximal}, we include a commonly used on-policy method for flat policies which in addition serves as the foundation for the DAC algorithm.

As demonstrated in Figure \ref{fig:openaigym}, \met~performs better than or commensurate to existing option learning algorithms such as DAC, IOPG and Option-Critic as well as PPO.
Training mixture policies (via RHPO \citep{wulfmeier2019regularized}) without temporal abstraction slightly reduces both performance and sample efficiency but still outperforms on-policy methods in many cases.
Here, enabling temporal abstraction (even without explicitly maximizing it) provides an inductive bias to reduce the search space for the high-level controller and can lead to more data-efficient learning, such that \met~even without constraints performs better than RHPO.

Finally, while less data-efficient than \met, even off-policy learning alone with flat Gaussian policies (here MPO~\citep{abdolmaleki2018maximum}) can outperform current on-policy option algorithms, for example in the HalfCheetah and Swimmer domains while otherwise at least performing on par.
This emphasizes the importance of a strong underlying policy optimization method.
 
Using the switch constraints for increasing temporal abstraction from Section \ref{sec:method} can provide minor benefits in some tasks but has overall only a minor effect on performance. We further investigate this setting in sequential transfer in Section \ref{sec:sequential}.
It has to be noted that given the comparable simplicity of these tasks, considerable performance gains are achieved with pure off-policy training. 
In the next section, we study more complex domains to isolate additional gains from action and temporal abstraction.

\subsection{Ablations: Action Abstraction and Temporal Abstraction}
\label{sec:multi}  

\begin{figure}[t]
	\centering
	\begin{tabular}{cc}
        \includegraphics[trim=1in 0.2in 0.5in 0, clip,height = 0.175\textwidth]{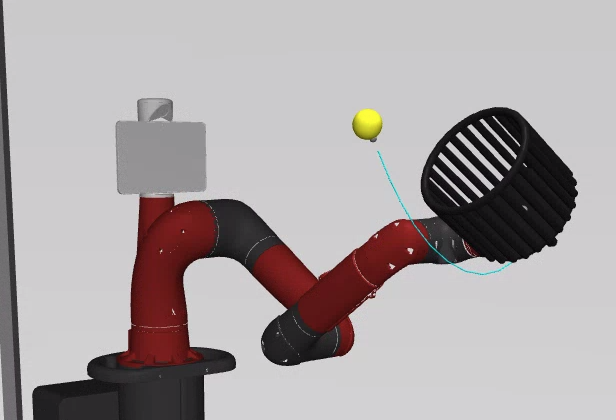}~
        \includegraphics[height = 0.175\textwidth]{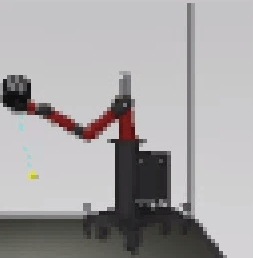}\\
        \includegraphics[trim=0.in 0.8in 0.in 0.in, clip,height = 0.19\textwidth]{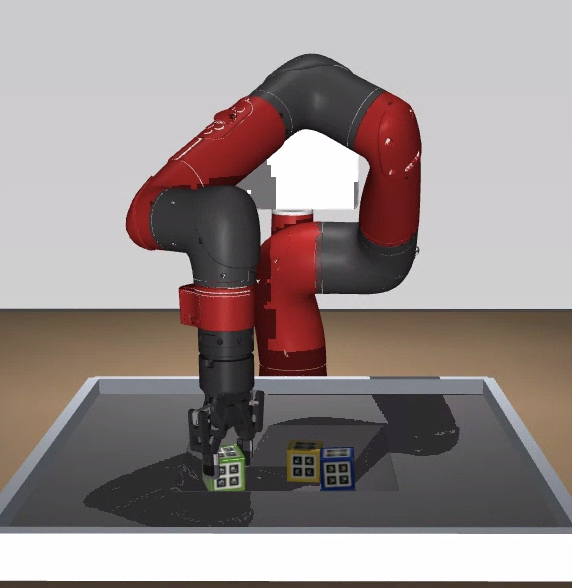}
        \includegraphics[height = 0.19\textwidth]{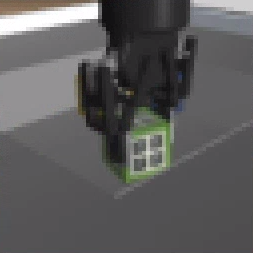}
	\end{tabular}
    \caption{Ball-In-Cup (top) and Stacking (bottom). Left: Environments. Right: Example agent observations.}
	\label{fig:evs}
\end{figure}

We next ablate different aspects of \met~on more complex simulated 3D robot manipulation tasks - stacking and the more dynamic ball-in-cup (BIC) - as displayed in Figure \ref{fig:evs}, based on robot proprioception and raw pixel inputs (64x64 pixel, 2 cameras for BIC and 3 for stacking). 
Since the performance of \met~for training from scratch is relatively independent of switch constraints (Figure \ref{fig:openaigym}), we will simplify our figures by focusing on the base method.
To reduce data requirements, we use a set of common techniques to improve data-efficiency and accelerate learning for all methods. We will apply multi-task learning with a related set of tasks to provide a curriculum,
with details in Appendix \ref{app:additiona_exps}. Furthermore, we assign rewards for all tasks to any generated transition data in hindsight to improve data efficiency and exploration \citep{andrychowicz2017hindsight,riedmiller2018learning,wulfmeier2019regularized, cabi2017intentional}.

\begin{figure}[t]
	\centering
	\setlength{\tabcolsep}{0pt}
	\begin{tabular}{cc}
        \includegraphics[width = 0.24\textwidth]{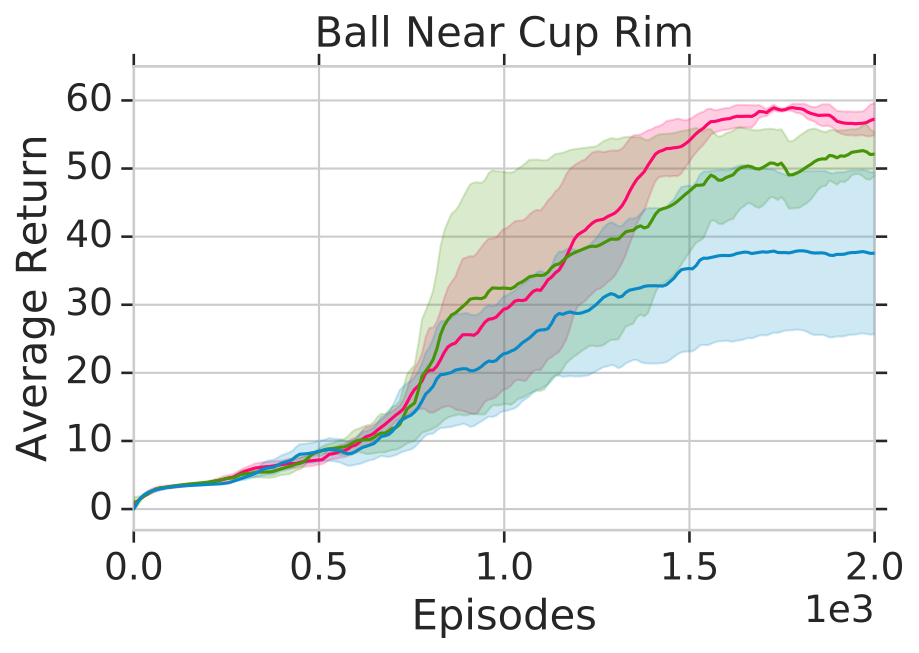}
        \includegraphics[width = 0.24\textwidth]{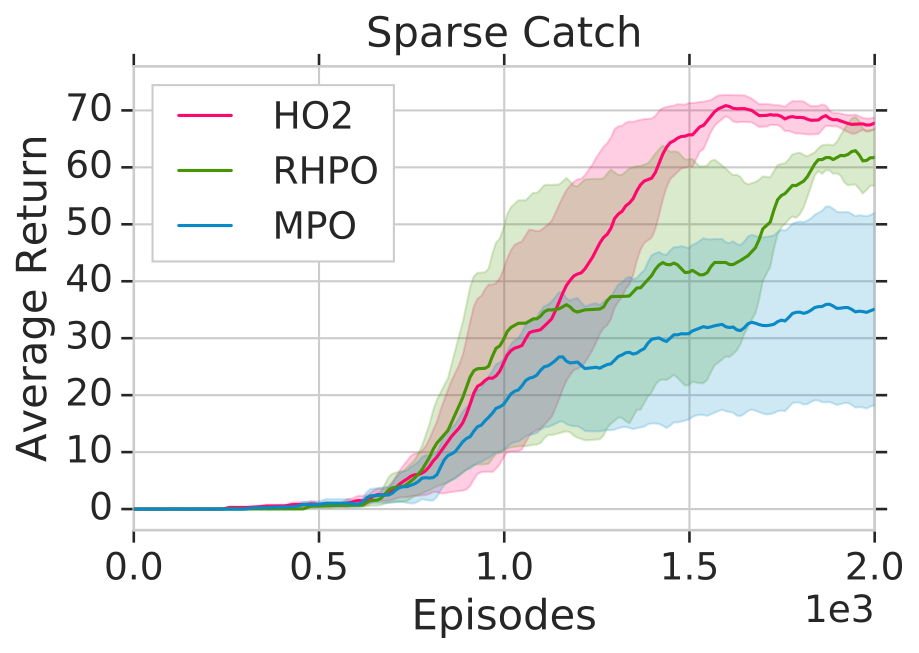}\\
        \includegraphics[width = 0.24\textwidth]{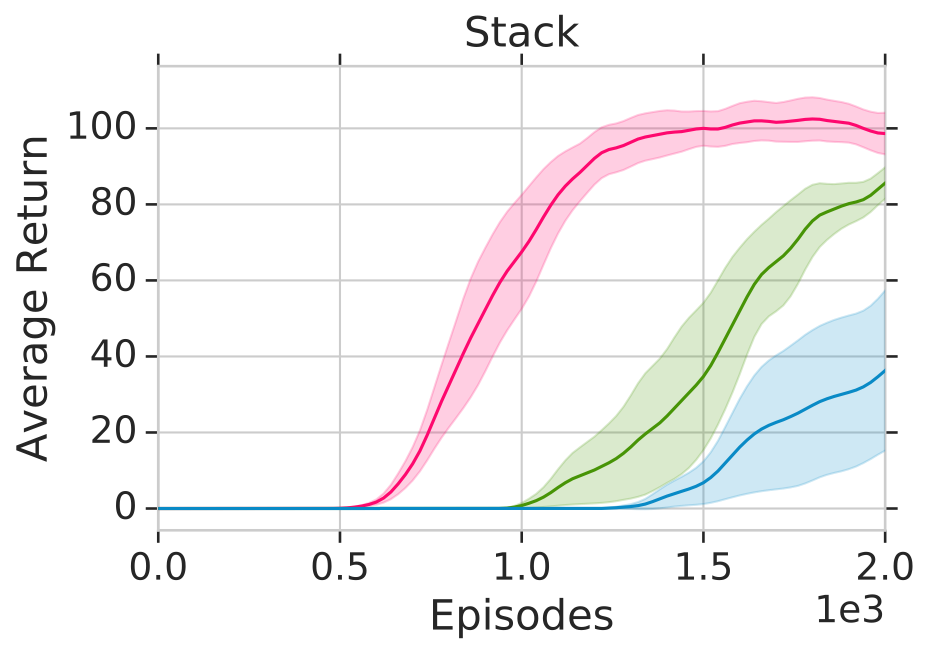}
        \includegraphics[width = 0.24\textwidth]{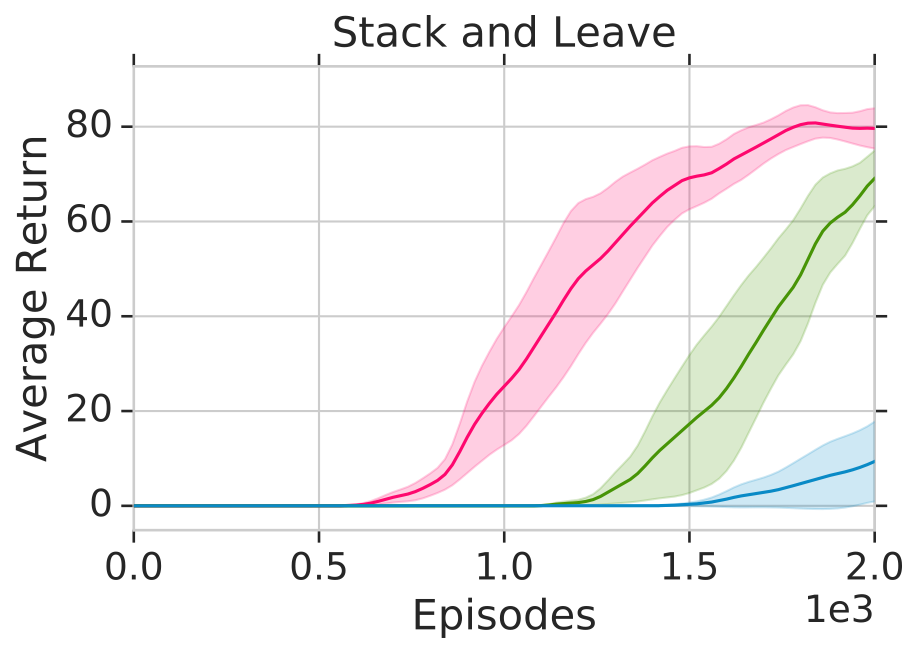}
	\end{tabular}
	\vspace{-3pt}
    \caption{Results for option policies, and ablations via mixture policies, and single Gaussian policies (respectively \met, RHPO, and MPO) with pixel-based ball-in-cup (left) and pixel-based block stacking (right). All four tasks displayed use sparse binary rewards, such that the obtained reward represents the number of timesteps where the corresponding condition - such as the ball is in the cup - is fulfilled. 
    See Appendix \ref{app:details} for details and additional tasks.}
	\label{fig:multitask}
	\vspace{-8pt}
\end{figure}

Across all tasks, except for simple positioning and reach tasks (see Appendix \ref{app:additiona_exps}), action abstraction improves performance (mixture policies via RHPO versus flat Gaussian policies via MPO). 
In particular the more challenging stacking tasks shown in Figure \ref{fig:multitask} intuitively benefit from shared sub-behaviors with easier tasks.
Finally, the introduction of temporal abstraction (option policies via \met~vs mixture policy via RHPO) further improves both performance and sample efficiency, especially on the more complex stacking tasks.
The ability to learn explicit termination conditions, which can be understood as classifiers between two conditions, instead of the high-level controller, as classifier between all options, can considerably simplify the learning problem.

\paragraph{Optimizing for Temporal Abstraction}
\label{sec:sequential}
There is a difference between simplifying the representation of temporal abstraction for the agent and explicitly maximizing it. 
The ability to represent temporally abstract behavior in \met~via the use of explicit termination conditions consistently helps in prior experiments. 
However, these experiments show limited benefit when increasing temporal consistency (by limiting the number of switches following Section \ref{sec:temporal_abstraction}) for training from scratch.

\begin{figure}[t]
	\centering
	\setlength{\tabcolsep}{0pt}
	\begin{tabular}{cc}
        \includegraphics[trim=3mm 0 0 0,clip,height = 0.158\textwidth]{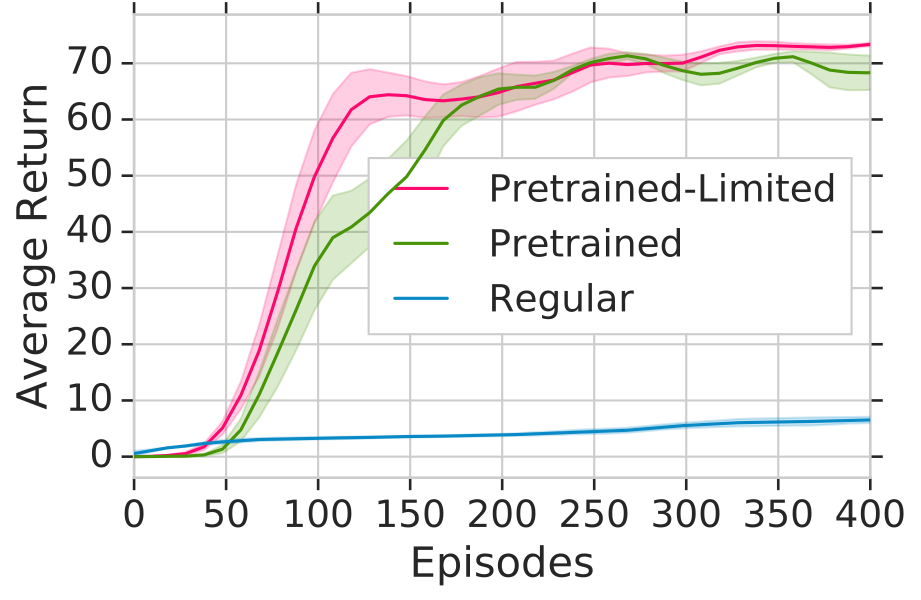} 
        \includegraphics[height = 0.158\textwidth]{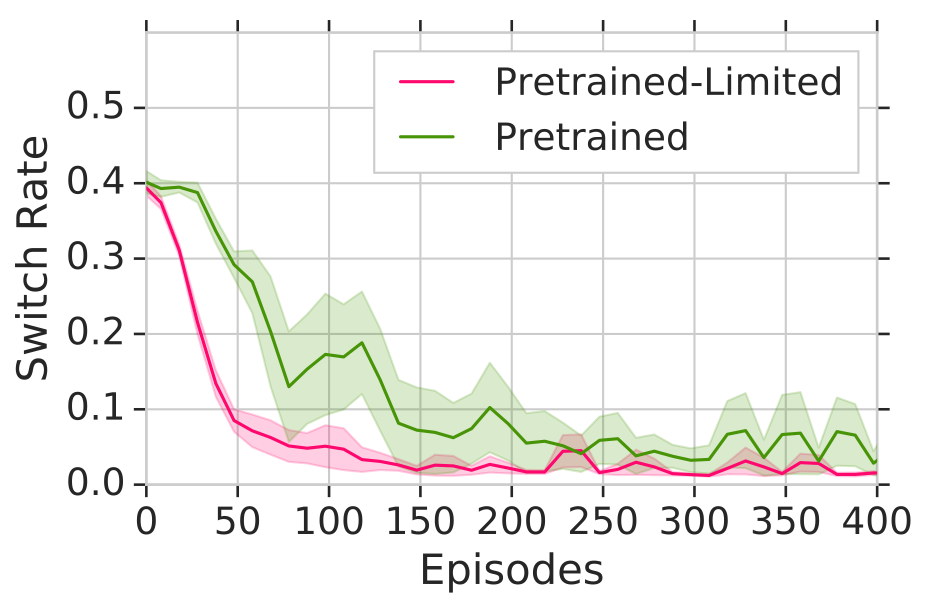}\\
        \includegraphics[height = 0.158\textwidth]{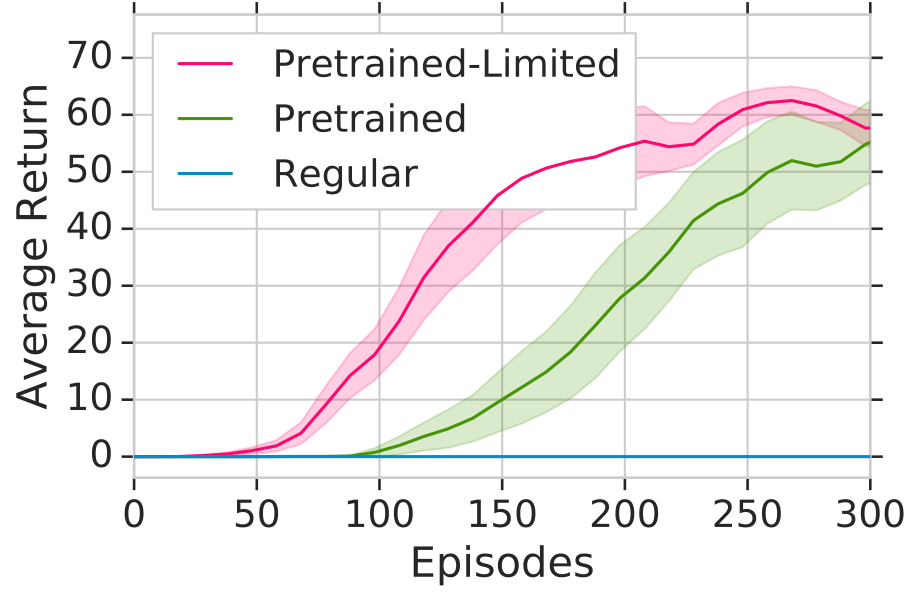} 
        \includegraphics[height = 0.158\textwidth]{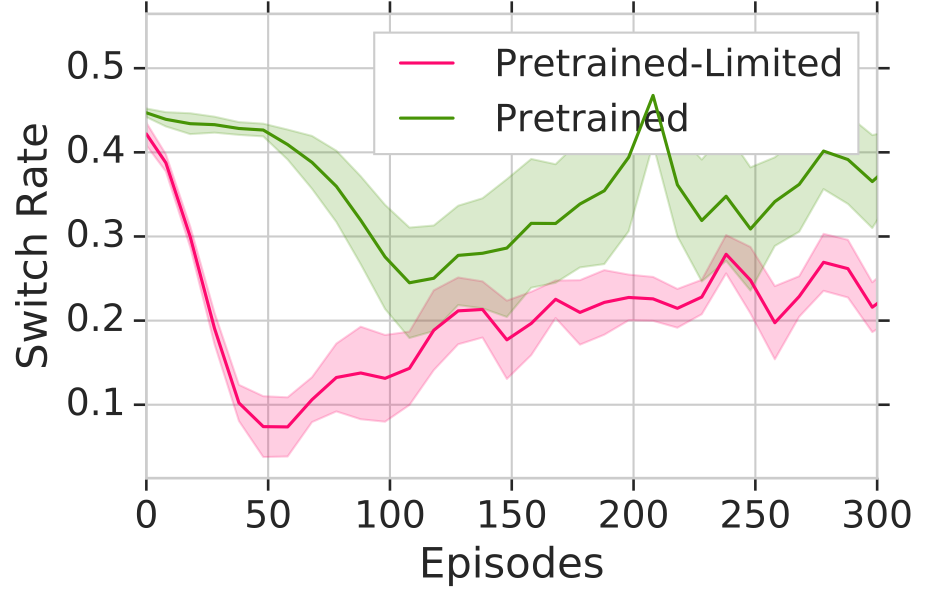}
	\end{tabular}
    \caption{The sequential transfer experiments for temporal abstraction show considerable improvements for limited switches. Top: BIC. Bottom: Stack.  In addition, we visualize the actual agent option switch rate in the environment to directly demonstrate the constraint's effect.}
	\label{fig:sequential}
\end{figure}

In this section, we further evaluate temporal abstraction for sequential transfer with pre-trained options.
We first train low-level options for all tasks except for the most complex task in each domain by applying \met. Next, given a set of pre-trained options, we only train the final task and compare training with and without maximizing temporal abstraction.  
We use the domains from Section \ref{sec:multi}, block stacking and BIC.
As shown in Figure \ref{fig:sequential}, we can see that more consistent options lead to increased performance in the transfer domain. Intuitively, increased temporal consistency and fewer switches lead to a smaller search space from the perspective of the high-level controller. 

While the same mechanism should also apply for training from scratch, we hypothesize that the added complexity of simultaneously learning the low-level behaviors (while maximizing temporal consistency) outweighs the benefits. Finding a set of options which not only solve a task but also represent temporally consistent behavior can be harder, and require more data, than just solving the task. 

\subsection{Ablations: Off-policy, Robustness, Decomposition} \label{sec:ablations}
In this section, we investigate different algorithmic aspects to get a better understanding of the method, properties of the learned options, and how to achieve robust training in the off-policy setting.

\paragraph{Off-policy Option Learning}

\begin{figure}[h]
	\centering
	\setlength{\tabcolsep}{1pt}
	\begin{tabular}{cc}
	\includegraphics[trim=.1in 0 17.5in 0,clip,height = 0.165\textwidth]{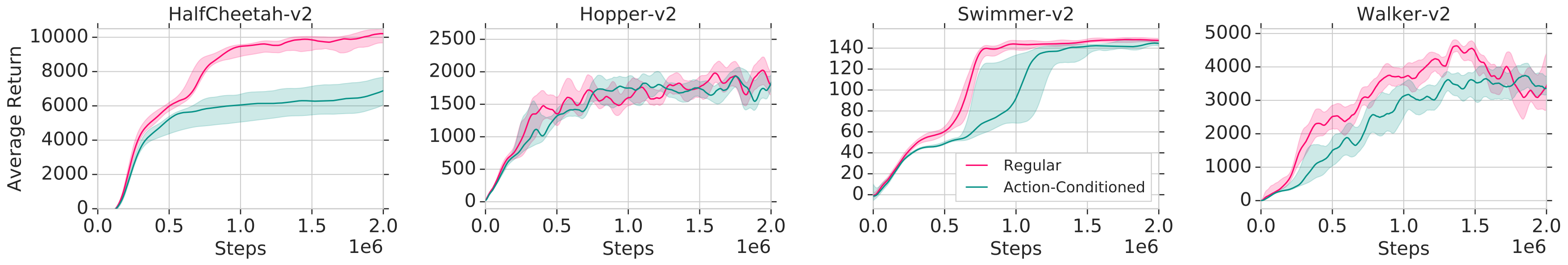} %
	\includegraphics[trim=12in 0 5.7in 0,clip,height = 0.165\textwidth]{results/png/gym_ac.png} %
	\end{tabular}
    \caption{Results on OpenAI gym with/without option probabilities being conditioned on past actions.
    }
	\label{fig:ac}
\end{figure}

In off-policy hierarchical RL, the low-level policy underlying an option can change after trajectories are generated. 
This results in a non-stationarity for training the high-level controller. 
In addition, including previously executed actions in the forward computation for component probabilities can introduce additional variance into the objective.
In practice, we find that removing the conditioning on low-level probabilities (the $\pi_L$ terms in Equation \ref{eq:dynamic1}) improves performance and stability.
The effect is displayed in Figure \ref{fig:ac}, where the conditioning of high-level component probabilities on
low-level action probabilities (see Section \ref{sec:method}) is detrimental.  

We additionally evaluate this effect in the on-policy setting in Appendix~\ref{sec:off_policy_exps} and find its impact to be diminished, demonstrating the connection between the effect and an off-policy setting. 
While we apply this simple heuristic for \met, the problem has lead to various off-policy corrections for goal-based HRL \citep{nachum2018data,levy2017learning} which provide a valuable direction for future work.

\paragraph{Trust-regions and Robustness}

\begin{figure}[h]
      \centering
      \begin{tabular}{cc}
        \includegraphics[trim=3mm 3mm 0 0,clip,width=0.23\textwidth]{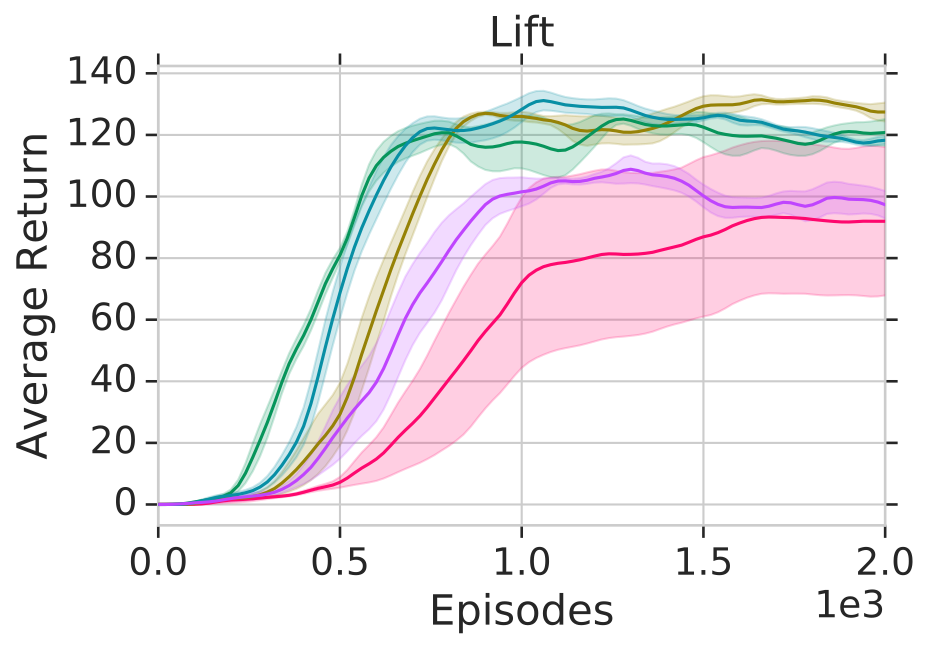}
        \includegraphics[trim=3mm 3mm 0 0,clip,width=0.23\textwidth]{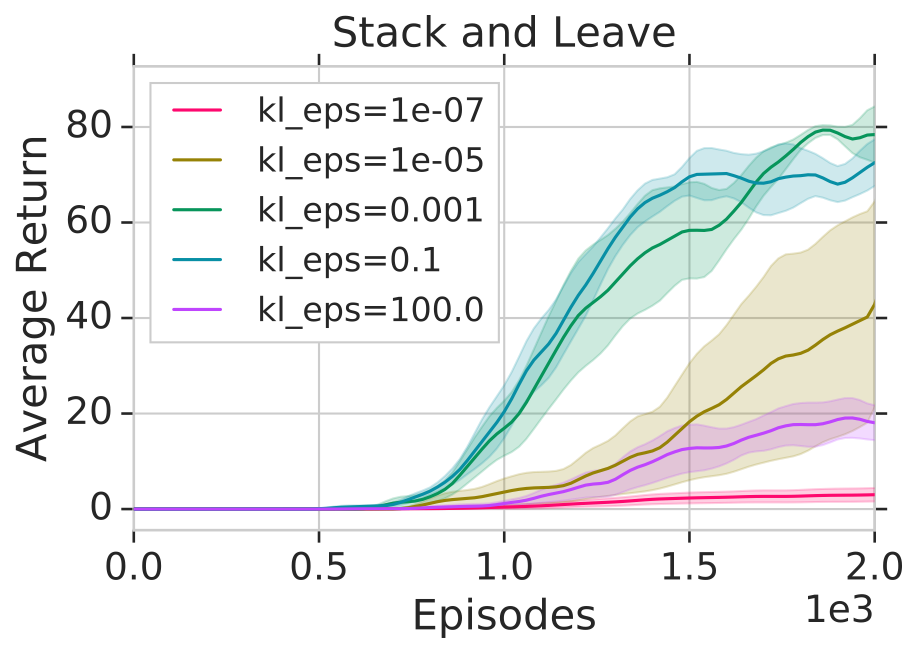}
      \end{tabular}
      \caption{Block stacking results for two tasks with different trust-region constraints. Note that the importance of constraints increases for more complex tasks.}
      \label{fig:trustregion}
\end{figure}

Previous work has shown the benefits of applying trust-region constraints for policy updates of non-hierarchical policies \citep{schulman2015trust, abdolmaleki2018maximum}. 
In this section, we vary the strength of constraints on the option probability updates (both termination conditions $\beta$ and the high-level controller $\pi_C$). 
As displayed in Figure~\ref{fig:trustregion}, the approach is robust across multiple orders of magnitude, but very 
weak or strong constraints can considerably degrade performance. 
Note that a high value is essentially equal to not using a constraint and causes very low performance. Therefore, option learning here relies strongly on trust-region constraints. Further experiments can be found in Appendix~\ref{sec:trust_region_exps}.

\begin{figure}[h]
        \includegraphics[scale = 0.19, trim=3mm 0 0 0, clip]{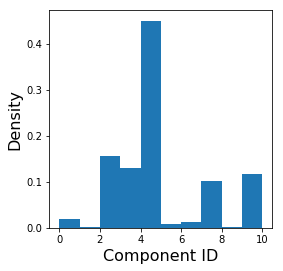}
        \includegraphics[scale = 0.21, trim=0 0 0 3mm, clip]{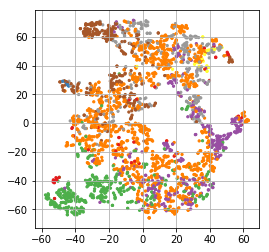}
        \includegraphics[scale = 0.19, trim=3mm 0 0 0, clip]{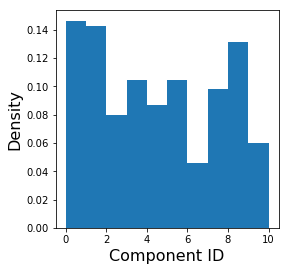}
        \includegraphics[scale = 0.21, trim=0 0 0 3mm, clip]{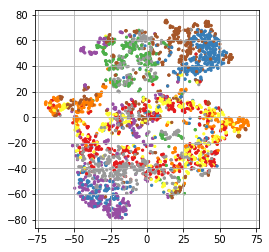}
    \caption{Analysis on Ant locomotion tasks, showing histogram over options, and t-SNE scatter plots in action space colored by option. Left: without IA. Right: with IA. Agents with IA use more components and show clearer option clustering in the action space.}
	\label{fig:predicate_decomp}
\end{figure}

\paragraph{Decomposition and Option Clustering}
To investigate how \met~uses its capacity and decomposes behavior into options, we apply it to a variety of simple and interpretable locomotion tasks.
In these tasks, the agent body (``Ball'', ``Ant'', or ``Quadruped'') must go to one of three targets in a room, with the task specified by the target locations and a selected target index.
As shown for the ``Ant'' case in Figure~\ref{fig:predicate_decomp}, we find that option decomposition intuitively depends on both the task properties and algorithm settings. In particular \textit{information asymmetry (IA)}, achieved by providing task information only to the high-level controller, can address degenerate solutions and lead to increased diversity with respect to options (as shown by the histogram over options) and more specialized options (represented by the clearer clustering of samples in action space).
We can measure this quantitatively, using (1) the Silhouette score, a measure of clustering accuracy based on inter- and intra-cluster distances\footnotemark; and (2) entropy over the option histogram, to quantify diversity.
These metrics are reported in Table~\ref{table:predicates_quant_main} for all bodies, with and without information asymmetry.
The results show that for all cases, IA leads to greater option diversity and clearer separation of option clusters with respect to action space, state space, and task. 

\footnotetext{The silhouette score is a value in $[-1, 1]$ with higher values indicating cluster separability. We note that the values obtained in this setting do not correspond to high \textit{absolute} separability, as multiple options can be used to model the same skill or behavior abstraction. We are instead interested in the \textit{relative} clustering score for different scenarios.}

More extensive experiments and discussion can be found in Appendix~\ref{sec:locomotion}, including additional quantitative and qualitative results for the other bodies and scenarios.
To summarize, the analyses yield a number of relevant observations, showing that (1) for simpler bodies like a ``Ball'', the options are interpretable (forward torque, and turning left/right at different rates); and (2) applying the switch constraint introduced in Section~\ref{sec:sequential} leads to increased temporal abstraction without reducing the agent's ability to solve the task.

\begin{table}[!h]
\centering
\scalebox{0.65}{
\begin{tabular}{ c c | c  c  c  c }
  \hline \\ [-1.5ex]
  \multicolumn{2}{c}{\bf{Scenario}} & \bf{Option entropy} & \bf{s (action)}  & \bf{s (state)}  & \bf{s (task)} \\
 \hline \\ [-1.5ex]
 \multirow{2}{*}{Ball} & No IA & $1.80 \pm 0.21$ & $-0.30 \pm 0.04$ & $-0.25 \pm 0.14$ & $-0.13 \pm 0.05$ \\
  & With IA & $2.23 \pm 0.03$ & $-0.13 \pm 0.04$ & $-0.11 \pm 0.04$ & $-0.05 \pm 0.00$ \\
 \cline{2-6} \\ [-1.5ex]
 \multirow{2}{*}{Ant} & No IA & $1.60 \pm 0.08$ & $-0.12 \pm 0.02$ & $-0.15 \pm 0.07$ & $-0.08 \pm 0.03$ \\
 & With IA & $2.22 \pm 0.04$ & $-0.05 \pm 0.01$ & $-0.05 \pm 0.01$ & $-0.05 \pm 0.01$ \\
 \cline{2-6} \\ [-1.5ex]
 \multirow{2}{*}{Quad} & No IA & $1.55 \pm 0.29$ & $-0.07 \pm 0.04$ & $-0.12 \pm 0.03$ & $-0.11 \pm 0.02$\\
 & With IA & $2.23 \pm 0.04$ & $-0.03 \pm 0.03$ & $-0.03 \pm 0.00$ & $-0.05 \pm 0.01$ \\
 \hline
\end{tabular}}
\vspace{1ex}
\caption{Quantitative results 
indicating the diversity of options used (entropy), and clustering accuracy in action and state spaces (silhouette score $s$), with and without information asymmetry (IA). Switching constraints are applied in all cases. 
Higher values indicate greater separability by option / component.}
\label{table:predicates_quant_main}
\end{table}

\section{Related Work}\label{sec:related}

Hierarchy has been investigated in many forms in reinforcement learning to improve data gathering as well as data fitting aspects. 
Goal-based approaches commonly define a grounded interface between high- and low-level policies.  The high level acts by providing goals to the low level, which is trained to achieve these goals \citep{dayan1993feudal,levy2017learning,nachum2018near,nachum2018data,vezhnevets2017feudal}, effectively generating separate objectives and improving exploration.
These methods have been able to overcome very sparse reward domains but commonly require domain knowledge to define the interface. In addition, a hand-crafted interface can limit expressiveness of achievable behaviors.

Non-crafted, emergent interfaces within policies have been investigated from an RL-as-inference perspective via policies with continuous latent variables \citep{haarnoja2018latent,hausman2018learning,heess2016learning,igl2019multitask,tirumala2019exploiting,teh2017distral}. Related to these approaches, we provide a probabilistic inference perspective to off-policy option learning and benefit from efficient dynamic programming inference procedures.
We furthermore build on the related idea of information asymmetry \citep{pinto2017asymmetric,galashov2018information,tirumala2019exploiting} - providing a part of the observations only to a part of the model. The asymmetry can lead to an information bottleneck affecting the properties of learned low-level policies. We build on the intuition and demonstrate how option diversity can be affected in ablations in Section \ref{sec:ablations}.

At its core, our work builds on and investigates the option framework \citep{precup2000temporal,sutton1999between},
which can be seen as describing policies with an autoregressive, discrete latent space.
Option policies commonly use a high-level controller to choose from a set of options or skills. These options additionally include termination conditions to enable a skill to represent temporally extended behavior. 
Without termination conditions, options can be seen as equivalent to components under a mixture distribution, and this simplified formulation has been applied successfully in different methods \citep{agostini2010reinforcement,daniel2016hierarchical,wulfmeier2019regularized}. 

Recent work has also investigated temporally extended low-level behaviours of fixed length  \citep{frans2018meta,li2019sub,nachum2018data}, which do not learn the option duration or termination condition.
With \met, enabling to optimize the extension of low-level behaviour in the option framework provides additional flexibility and removes the engineering effort of choosing the right hyperparameters.

The option framework has been further extended and improved for more practical application \citep{bacon2017option,harb2018waiting,harut2019termination,NIPS2005_2767,riemer2018learning,smith2018inference}.
\met~ relies on off-policy training and treats options as latent variables. This enables backpropagation through the option inference procedure and considerable improvements in comparison to efficient than approaches relying on on-policy updates and on-option learning purely for executed options. 
Related, IOPG \citep{smith2018inference} also considers an inference perspective but only includes on-policy results which naturally have poorer data efficiency.
Finally, the benefits of options and other modular policy styles have also been applied in the supervised case for learning from demonstration \citep{fox2017multi, krishnan2017ddco,shiarlis2018taco}.

One important step to increase the robustness of option learning has been taken in \citep{zhang2019dac} by building on robust (on-policy) policy optimization with PPO \citep{schulman2017proximal}. \met~has similar robustness benefits, but additionally improves data-efficiency by building on off-policy learning, hindsight inference of options, and additional trust-region constraints \citep{abdolmaleki2018maximum,wulfmeier2019regularized}. Related inference procedures have also been investigated in imitation learning \citep{shiarlis2018taco} as well as on-policy RL \citep{smith2018inference}.

In addition to inferring options in hindsight, off-policy learning enables us to assign rewards for multiple tasks, which has been successfully applied with flat, non-hierarchical policies \citep{andrychowicz2017hindsight,riedmiller2018learning, cabi2017intentional} and goal-based hierarchical approaches \citep{levy2017learning,nachum2018data}.

\section{Conclusions}\label{sec:conclusions}
We introduce a robust, efficient algorithm for off-policy training of option policies. The approach outperforms recent work in option learning on common benchmarks and is able to solve complex, simulated robot manipulation tasks from raw pixel inputs more reliably than competitive baselines.
\met~takes a probabilistic inference perspective to option learning, infers option and action probabilities for trajectories in hindsight, and performs critic-weighted maximum-likelihood estimation by backpropagating through the inference step.
Being able to infer options for a given trajectory allows robust off-policy training and determination of updates for all instead of only for the executed options. It also makes it possible to impose constraints on the termination frequency independently of an environment's reward scale. 

We separately analyze the impact of action abstraction (via mixture policies), and temporal abstraction (via options). We find that each abstraction independently improves performance. Additional maximization of temporal consistency for option choices is beneficial when transferring pre-trained options but displays a limited effect when learning from scratch.
Furthermore, we investigate the consequences of the off-policyness of training data and demonstrate the benefits of trust-region constraints for option learning.
We examine the impact of different agent and environment properties (such as information asymmetry, tasks, and embodiments) with respect to task decomposition and option clustering; a direction which provides opportunities for further investigation in the future.
Finally, since our method is based on (weighted) maximum likelihood estimation, it can be adapted naturally to learn structured behavior representations in mixed data regimes, e.g. to learn from combinations of demonstrations, logged data, and online trajectories. This opens up promising directions for future work.

\section*{Acknowledgments}
The authors would like to thank Peter Humphreys, Satinder Baveja, Tobias Springenberg, and Yusuf Aytar for helpful discussion and relevant feedback which helped to shape the publication.
We additionally like to acknowledge the support of the DeepMind robotics lab for infrastructure and engineering support.

\clearpage

\bibliography{references}
\bibliographystyle{icml2021}

\clearpage
\appendix
{\Large \bf Supplementary Material}

\section{Additional Experiments \label{app:additiona_exps}}

\subsection{Decomposition and Option Clustering}
\label{sec:locomotion}
We further deploy \met~on a set of simple locomotion tasks, where the goal is for an agent to move to one of three randomized target locations in a square room.
These are specified as a set of target locations and a task index to select the target of interest.

The main research questions we aim to answer (both qualitatively and quantitatively) are: (1) How do the discovered options specialize and represent different behaviors?; and (2) How is this decomposition affected by variations in the task, embodiment, or 
algorithmic properties of the agent?
To answer these questions, we investigate a number of variations:
\begin{itemize}
\item Three bodies: a quadruped with two (``Ant'') or three (``Quad'') torque-controlled joints on each leg, and a rolling ball (``Ball'') controlled by applying yaw and forward roll torques.
\item With or without \textit{information asymmetry} (IA) between high- and low-level controllers, where the task index and target positions are withheld from the options and only provided to the categorical option controller.
\item With or without a limited number of switches in the optimization.
\end{itemize}
Information-asymmetry (IA) in particular, has recently been shown to be effective for learning general skills~\citep{galashov2018information}: by withholding task-information from low-level options, they can learn task-agnostic, temporally-consistent behaviors that can be composed by the option controller to solve a task.
This mirrors the setup in the aforementioned Sawyer tasks, where the task information is only fed to the high-level controller.

For each of the different cases, we qualitatively evaluate the trained agent over $100$ episodes, and generate histograms over the different options used, and scatter plots to indicate how options cluster the state/action spaces and task information.
We also present quantitative measures (over $5$ seeds) to accompany these plots, in the form of (1) Silhouette score, a measure of clustering accuracy based on inter- and intra-cluster distances\footnotemark; and (2) entropy over the option histogram, to quantify diversity.
The quantitative results are shown in Table~\ref{table:predicates_quant}, and the qualitative plots are shown in Figure~\ref{fig:predicates_qual}.
Details and images of the environment are in Section~\ref{sec:locomotion_details}.

\footnotetext{The silhouette score is a value in $[-1, 1]$ with higher values indicating cluster separability. We note that the values obtained in this setting do not correspond to high \textit{absolute} separability, as multiple options can be used to model the same skill or behavior abstraction. We are instead interested in the \textit{relative} clustering score for different scenarios.}

\begin{table*}[h!]
\centering
\scalebox{0.95}{
\begin{tabular}{ c | c | c c c}
    & \multirow{2}{*}{\bf{Histogram over options}} & \multicolumn{3}{c}{\bf{t-SNE scatter plots}} \\
    &  & \bf{Actions} & \bf{States} & \bf{Task} \\
    \hline \\ [-1.5ex]
    \rotatebox{90}{\bf{Ball, no IA}} & \includegraphics[scale = 0.19]{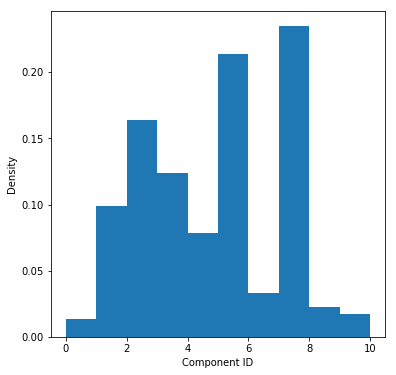}
    & \includegraphics[scale = 0.19]{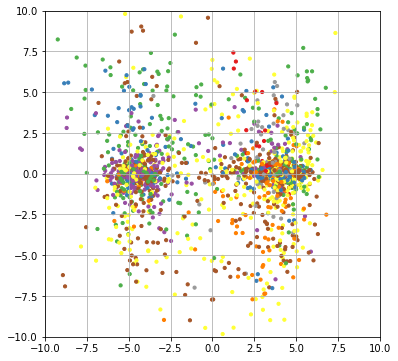} 
    & \includegraphics[scale = 0.19]{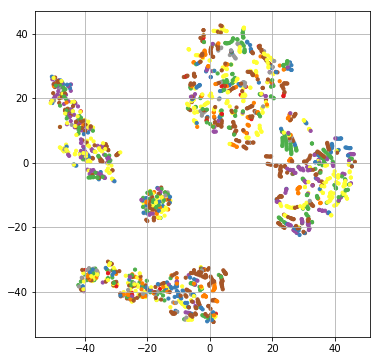} 
    & \includegraphics[scale = 0.19]{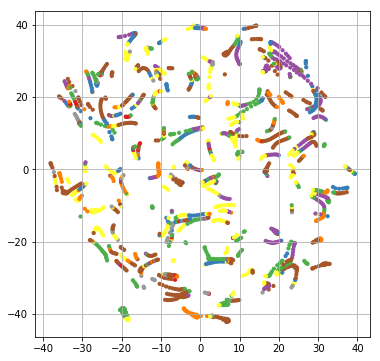} \\
    \rotatebox{90}{\bf{Ball, with IA}} & \includegraphics[scale = 0.19]{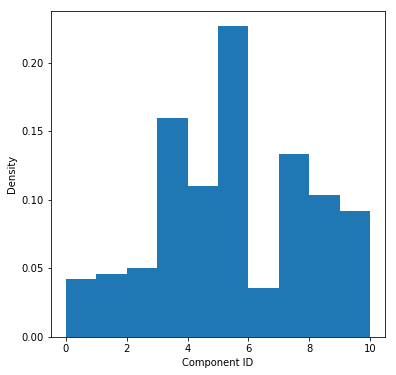}
    & \includegraphics[scale = 0.19]{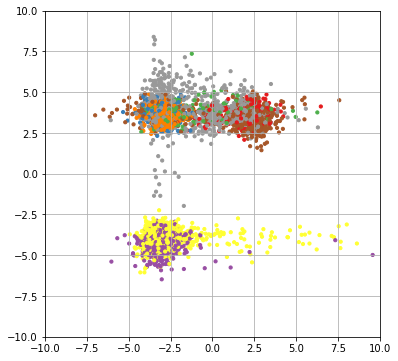} 
    & \includegraphics[scale = 0.19]{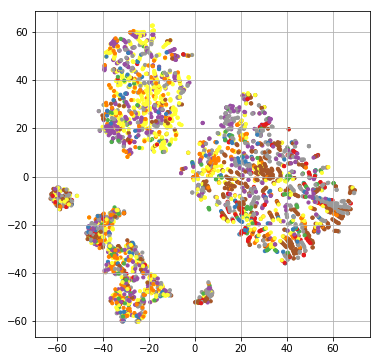} 
    & \includegraphics[scale = 0.19]{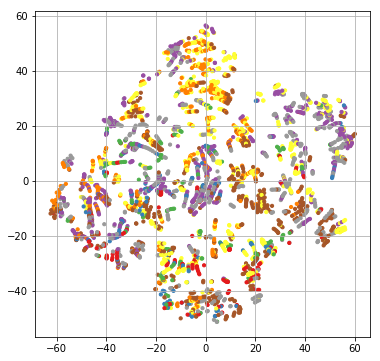} \\
    \rotatebox{90}{\bf{Ant, no IA}} & \includegraphics[scale = 0.19]{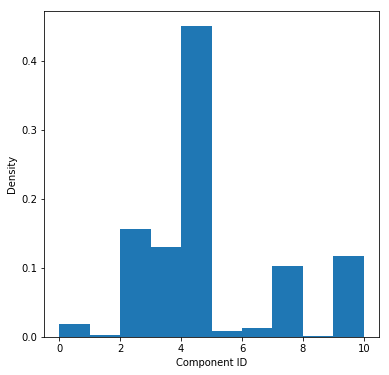}
    & \includegraphics[scale = 0.19]{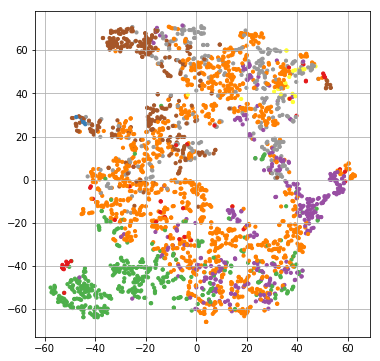} 
    & \includegraphics[scale = 0.19]{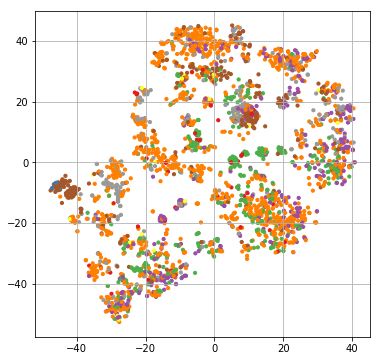}
    & \includegraphics[scale = 0.19]{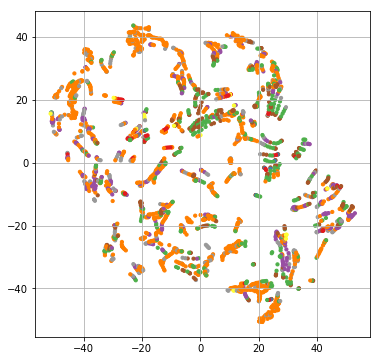} \\
    \rotatebox{90}{\bf{Ant, with IA}} & \includegraphics[scale = 0.19]{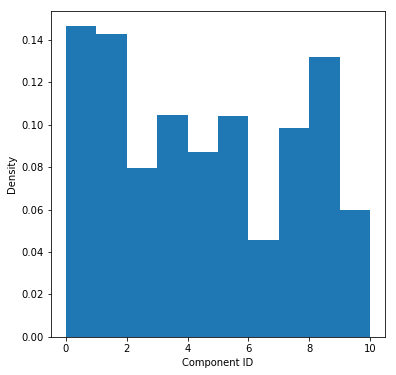}
    & \includegraphics[scale = 0.19]{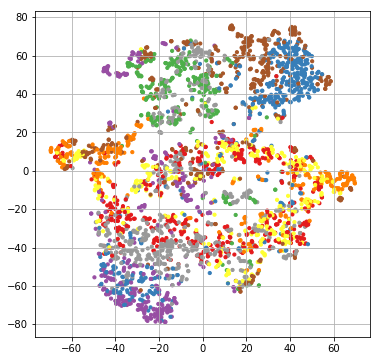} 
    & \includegraphics[scale = 0.19]{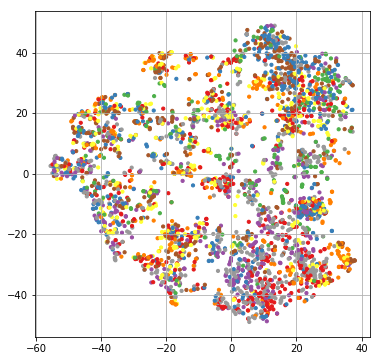} 
    & \includegraphics[scale = 0.19]{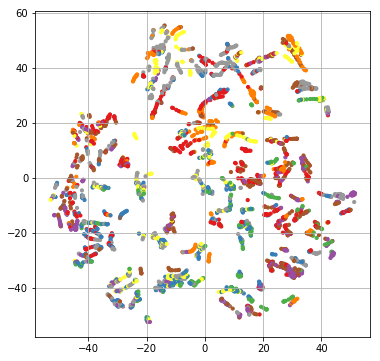} \\
    \rotatebox{90}{\bf{Quad, no IA}} & \includegraphics[scale = 0.19]{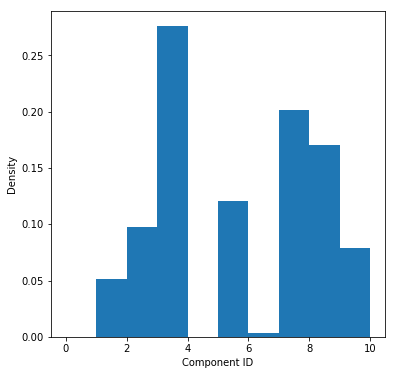}
    & \includegraphics[scale = 0.19]{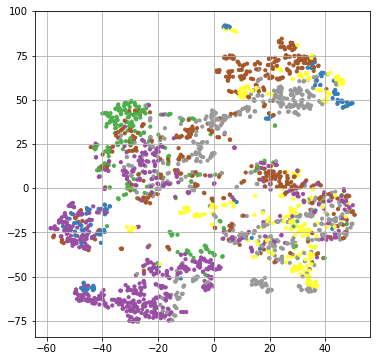} 
    & \includegraphics[scale = 0.19]{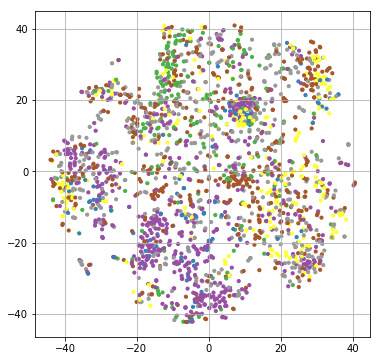} 
    & \includegraphics[scale = 0.19]{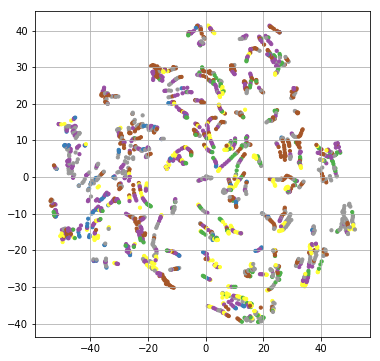} \\
    \rotatebox{90}{\bf{Quad, with IA}} & \includegraphics[scale = 0.19]{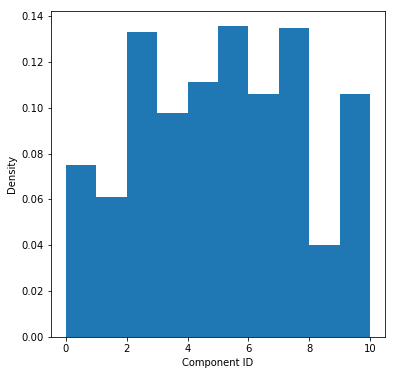}
    & \includegraphics[scale = 0.19]{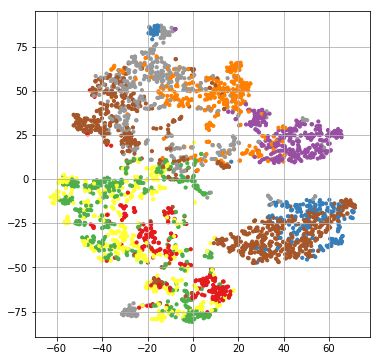} 
    & \includegraphics[scale = 0.19]{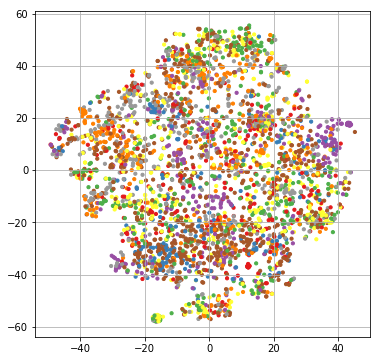} 
    & \includegraphics[scale = 0.19]{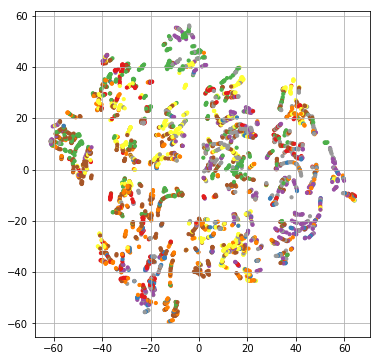} \\
    \vspace{1ex}
\end{tabular}}
\captionof{figure}{Qualitative results for the three bodies (Ball, Ant, Quad) without limited switches, both with and without IA, obtained over $100$ evaluation episodes. \textbf{Left}: the histogram over different options used by each agent; \textbf{Centre to right}: scatter plots of the action space, state space, and task information, colored by the corresponding option selected. Each of these spaces has been projected to $2D$ using t-SNE, except for the two-dimensional action space for Ball, which is plotted directly. For each case, care has been taken to choose a median / representative model out of $5$ seeds. }
\label{fig:predicates_qual}

\end{table*}

\begin{table*}[h]
\centering
\scalebox{0.7}{
\begin{tabular}{ c | c c | c  c  c  c  c }
  \hline \\ [-1.5ex]
  & \multicolumn{2}{c}{\bf{Scenario}} & \bf{Option entropy} & \bf{Switch rate} & \bf{Cluster score (actions)}  & \bf{Cluster score (states)}  & \bf{Cluster score (tasks)} \\
 \hline \\ [-1.5ex]
 \multirow{6}{*}{\rotatebox[origin=c]{90}{\bf{Regular}}} & \multirow{2}{*}{Ball} & No IA & $2.105 \pm 0.074$ & $0.196 \pm 0.010$ & $-0.269 \pm 0.058$ & $-0.110 \pm 0.025$ & $-0.056 \pm 0.011$ \\
  & & With IA & $2.123 \pm 0.066$ & $0.346 \pm 0.024$ & $-0.056 \pm 0.024$ & $-0.164 \pm 0.051$ & $-0.057 \pm 0.008$ \\
 \cline{4-8} \\ [-1.5ex]
 & \multirow{2}{*}{Ant} & No IA & $1.583 \pm 0.277$ & $0.268 \pm 0.043$ & $-0.148 \pm 0.034$ & $-0.182 \pm 0.068$ & $-0.075 \pm 0.011$ \\
 & & With IA & $2.119 \pm 0.073$ & $0.303 \pm 0.019$ & $-0.053 \pm 0.021$ & $-0.066 \pm 0.024$ & $-0.052 \pm 0.006$ \\
 \cline{4-8} \\ [-1.5ex]
 & \multirow{2}{*}{Quad} & No IA & $1.792 \pm 0.127$ & $0.336 \pm 0.019$ & $-0.078 \pm 0.064$ & $-0.113 \pm 0.035$ & $-0.089 \pm 0.050$ \\
 & & With IA & $2.210 \pm 0.037$ & $0.403 \pm 0.014$ & $0.029 \pm 0.029$ & $-0.040 \pm 0.003$ & $-0.047 \pm 0.006$ \\
 \hline \\ [-1.5ex]
 \multirow{6}{*}{\rotatebox[origin=c]{90}{\parbox{2cm}{\bf{Limited Switches}}}} & \multirow{2}{*}{Ball} & No IA & $1.804 \pm 0.214$ & $0.020 \pm 0.009$ & $-0.304 \pm 0.040$ & $-0.250 \pm 0.135$ & $-0.131 \pm 0.049$ \\
  & & With IA & $2.233 \pm 0.027$ & $0.142 \pm 0.015$ & $-0.132 \pm 0.035$ & $-0.113 \pm 0.043$ & $-0.053 \pm 0.003$ \\
 \cline{4-8} \\ [-1.5ex]
 & \multirow{2}{*}{Ant} & No IA & $1.600 \pm 0.076$ & $0.073 \pm 0.014$ & $-0.124 \pm 0.017$ & $-0.155 \pm 0.067$ & $-0.084 \pm 0.034$ \\
 & & With IA & $2.222 \pm 0.043$ & $0.141 \pm 0.015$ & $-0.052 \pm 0.011$ & $-0.054 \pm 0.014$ & $-0.050 \pm 0.007$ \\
 \cline{4-8} \\ [-1.5ex]
 & \multirow{2}{*}{Quad} & No IA & $1.549 \pm 0.293$ & $0.185 \pm 0.029$ & $-0.075 \pm 0.036$ & $-0.126 \pm 0.030$ & $-0.112 \pm 0.022$\\
 & & With IA & $2.231 \pm 0.042$ & $0.167 \pm 0.025$ & $-0.029 \pm 0.029$ & $-0.032 \pm 0.004$ & $-0.053 \pm 0.009$ \\
 \hline
\end{tabular}}
\vspace{1ex}
\caption{Quantitative results 
indicating the diversity of options used (entropy), and clustering accuracy in action and state spaces (silhouette score), with and without information asymmetry (IA), and with or without limited number of switches. 
Higher values indicate greater separability by option / component.}
\label{table:predicates_quant}
\end{table*}

The results show a number of trends.
Firstly, the usage of IA leads to a greater diversity of options used, across all bodies.
Secondly, with IA, the options tend to lead to specialized actions, as demonstrated by the clearer option separation in action space.
In the case of the $2D$ action space of the ball, the options correspond to turning left or right (y-axis) at different forward torques (x-axis).
Thirdly, while the simple Ball can learn these high-level body-agnostic behaviors, the options for more complex bodies have greater switch rates that suggest the learned behaviors may be related to lower-level motor behaviors over a shorter timescale.
Lastly, limiting the number of switches during marginalization does indeed lead to a lower switch rate between options, without hampering the ability of the agent to complete the task.

\subsection{Action and Temporal Abstraction Experiments}
\begin{figure*}[h]
	\centering
	\begin{tabular}{ccc}
        \includegraphics[width = 0.3\textwidth]{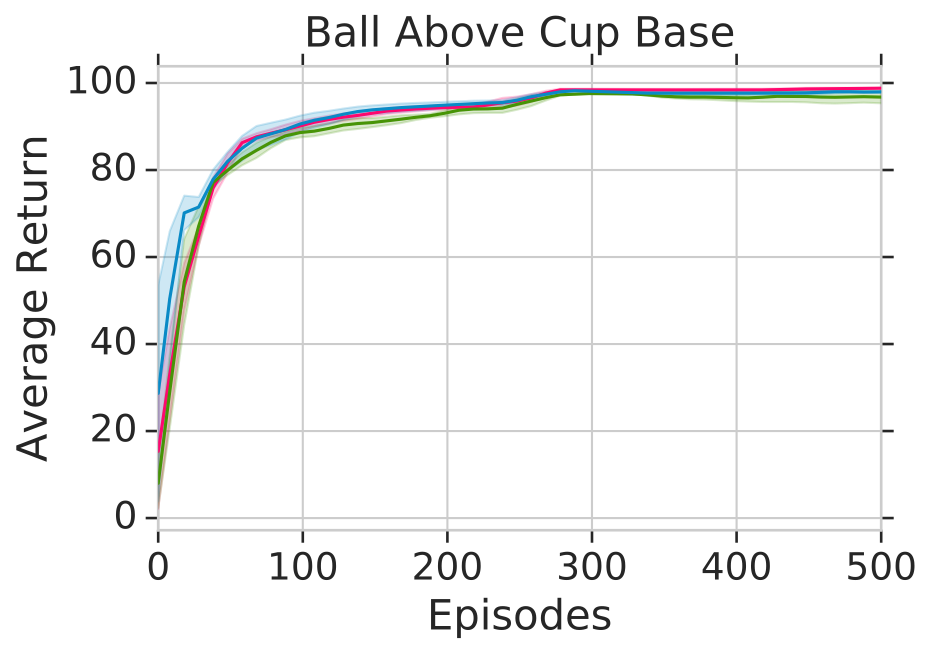}
        \includegraphics[width = 0.3\textwidth]{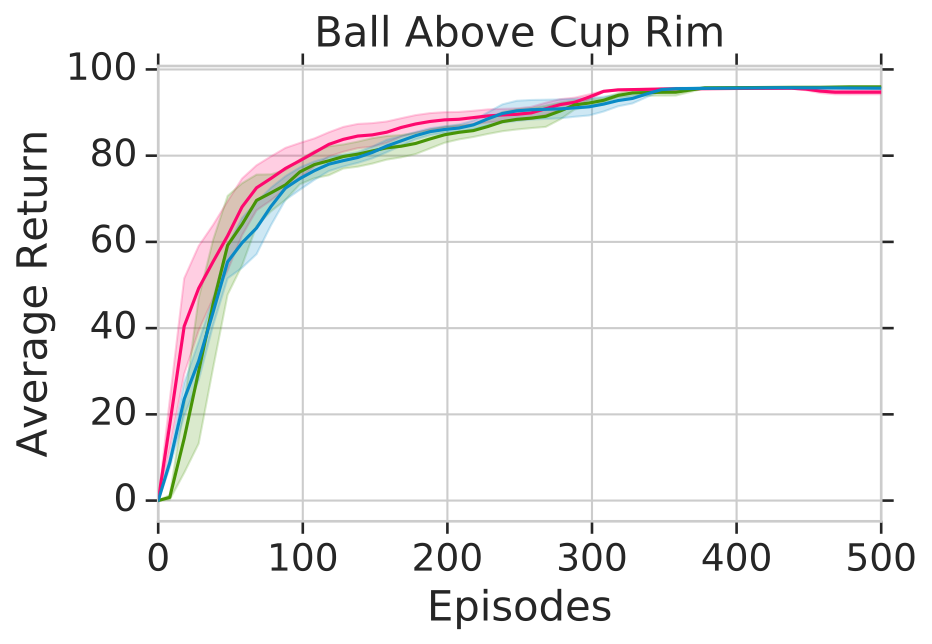}
        \includegraphics[width = 0.3\textwidth]{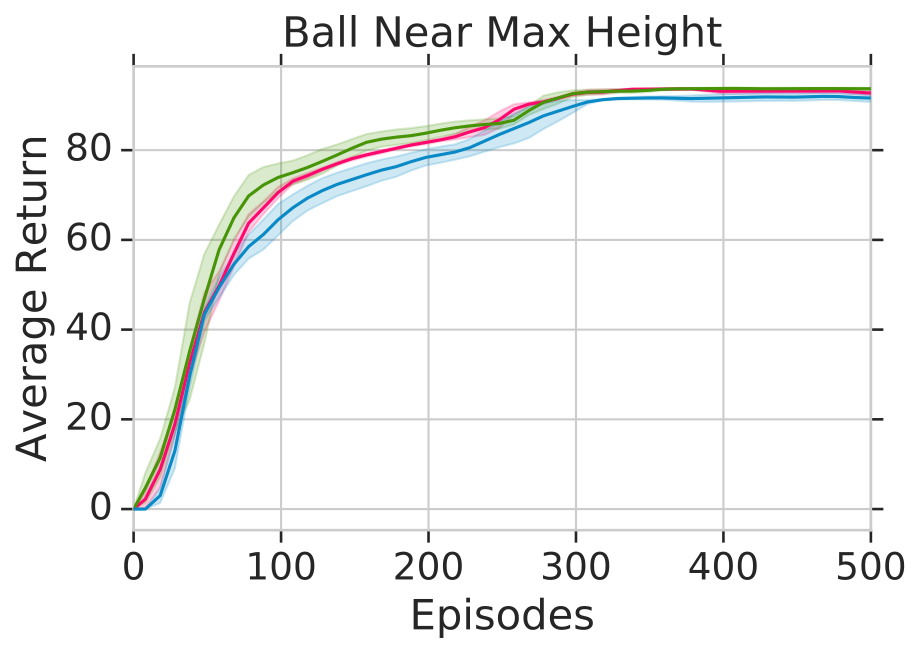}\\ 
        \includegraphics[width = 0.3\textwidth]{results/png/bic/BIC_4.png}
        \includegraphics[width = 0.3\textwidth]{results/png/bic/BIC_0.png} 
	\end{tabular}
    \caption{Complete results on pixel-based ball-in-cup experiments.}
	\label{fig:bic_all}
\end{figure*}

\begin{figure*}[h]
	\centering
	\begin{tabular}{cccc}
        \includegraphics[width = 0.235\textwidth]{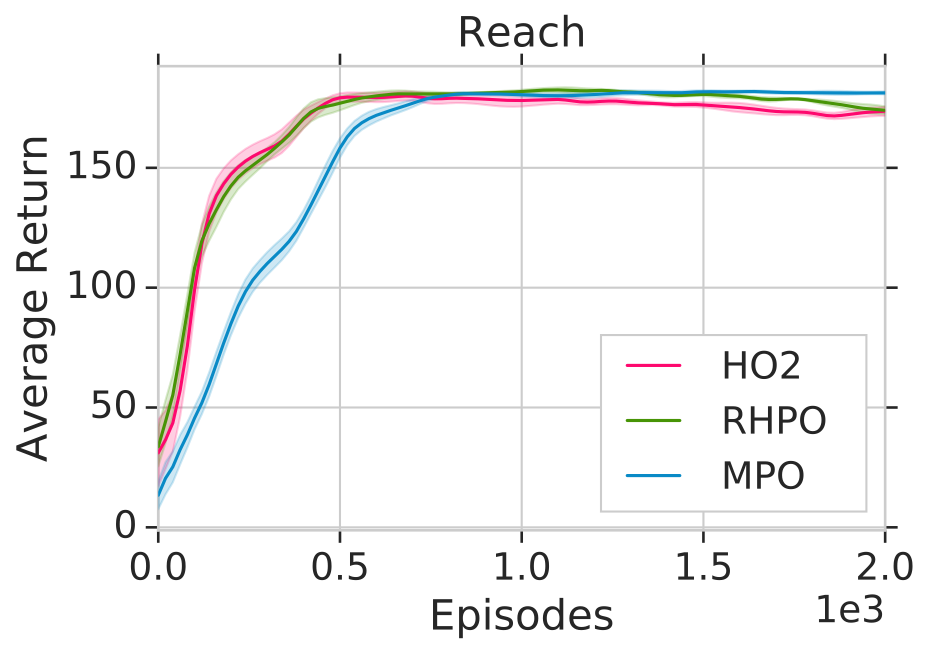} 
        \includegraphics[width = 0.235\textwidth]{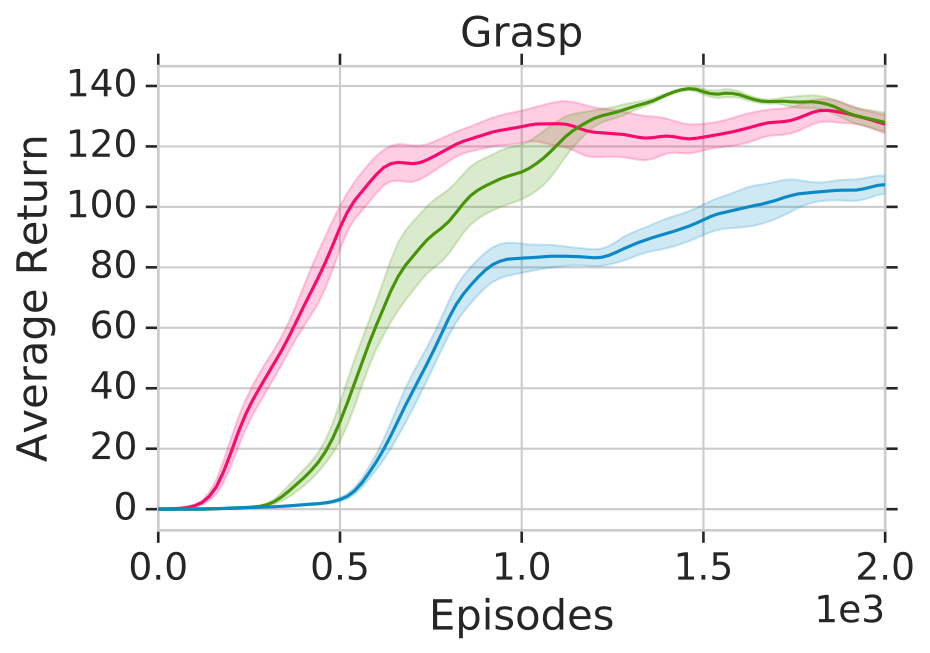} 
        \includegraphics[width = 0.235\textwidth]{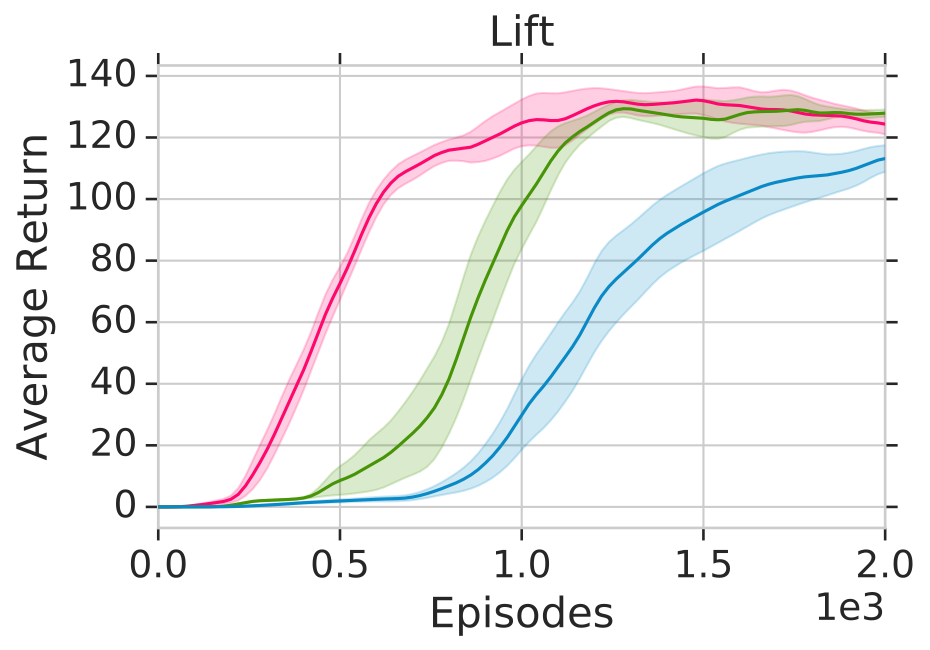} 
        \includegraphics[width = 0.235\textwidth]{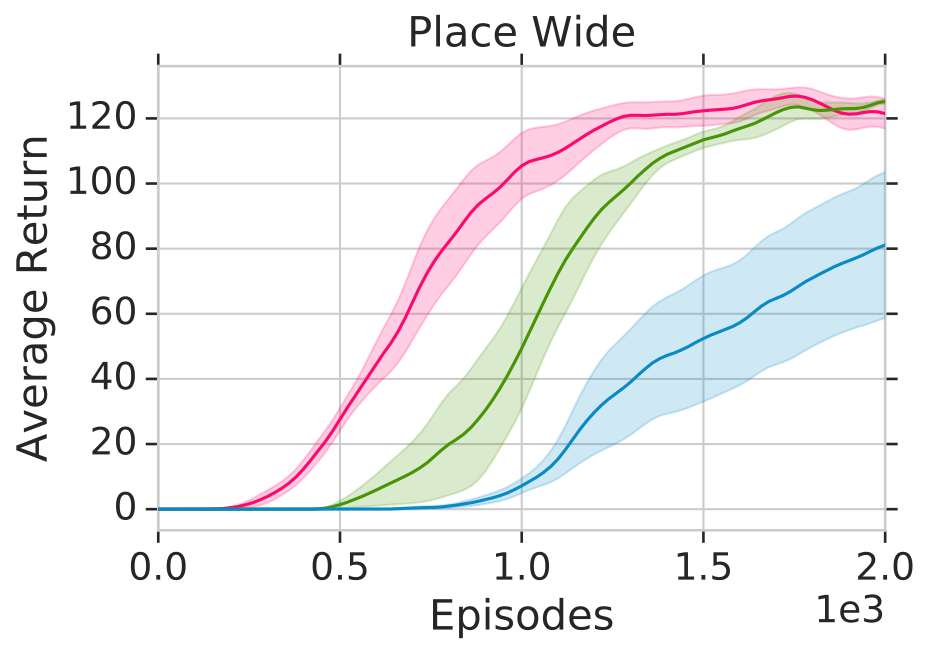} \\
        \includegraphics[width = 0.235\textwidth]{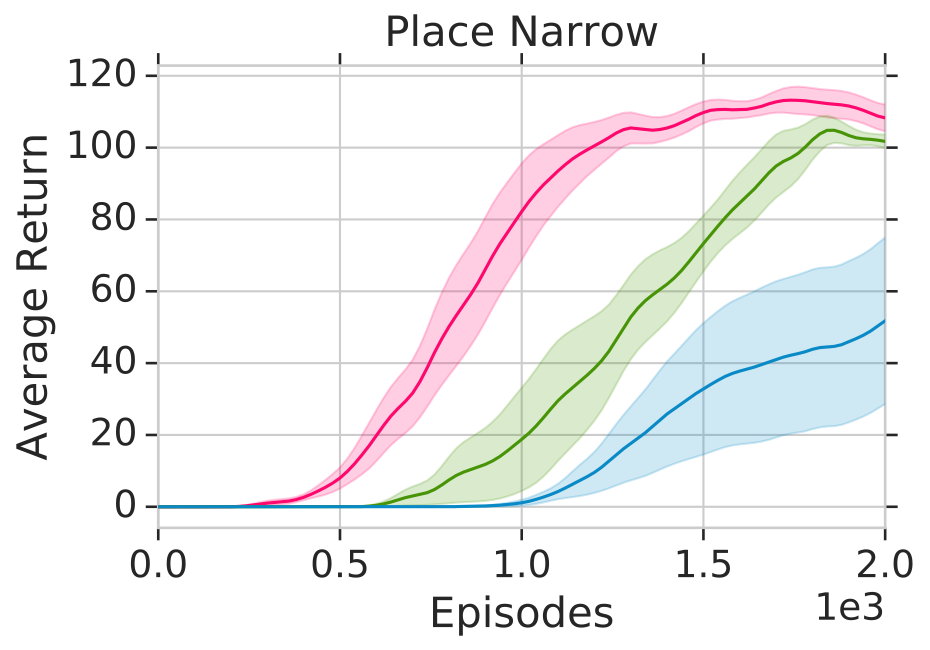} 
        \includegraphics[width = 0.235\textwidth]{results/png/stack/pile1_5.png} 
        \includegraphics[width = 0.235\textwidth]{results/png/stack/pile1_6.png} 
	\end{tabular}
    \caption{Complete results on pixel-based stacking experiments.}
	\label{fig:stack_all}
\end{figure*}

The complete results for all pixel and proprioception based multitask experiments for ball-in-cup and stacking can be respectively found in Figures \ref{fig:bic_all} and \ref{fig:stack_all}.
Both RHPO and \met~outperform a simple Gaussian policy trained via MPO.  \met~additionally improves performance over mixture policies (RHPO) demonstrating that the ability to learn temporal abstraction proves beneficial in these domains. The difference grows as the task complexity increases and is particularly pronounced for the final stacking tasks.

\subsection{Task-agnostic Terminations}
The perspective of options as task-independent skills with termination conditions as being part of a skill, leads to termination conditions which are also task independent. We show that at least in this limited set of experiments, the perspective of task-dependent termination conditions - i.e. with access to task information - which can be understood as part of the high-level control mechanism for activating options improves performance. 
Intuitively, by removing task information from the termination conditions, we constrain the space of solutions which first accelerates training slightly but limits final performance. It additionally shows that while we benefit when sharing options across tasks, each task gains from controlling the length of these options independently. 
Based on these results, the termination conditions across all other multi-task experiments are conditioned on the active task.

\begin{figure*}[h]
	\centering
	\begin{tabular}{cccc}
        \includegraphics[width = .235\textwidth]{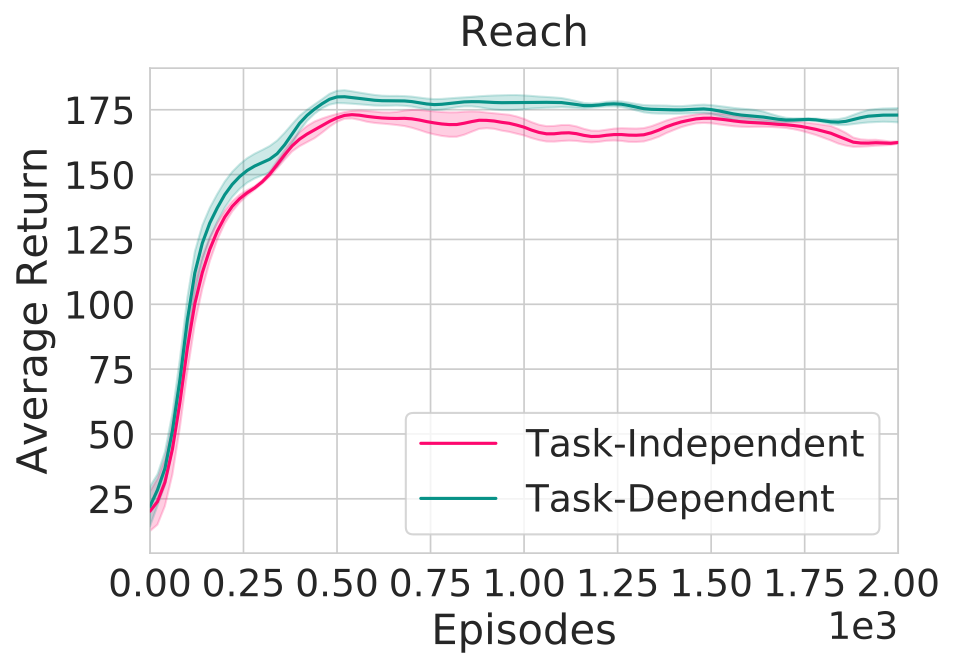} 
        \includegraphics[width = .235\textwidth]{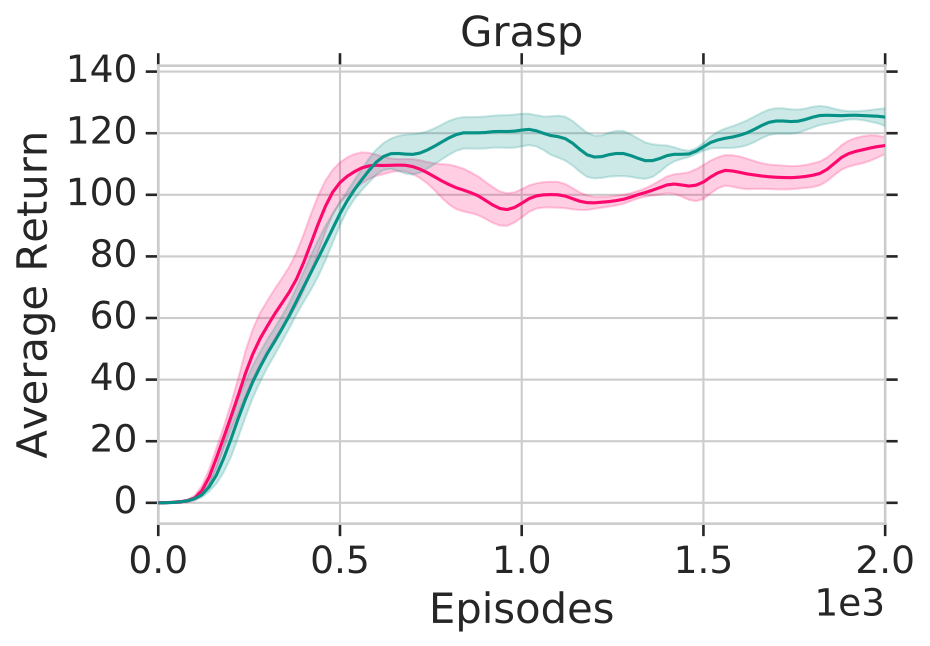}
        \includegraphics[width = .235\textwidth]{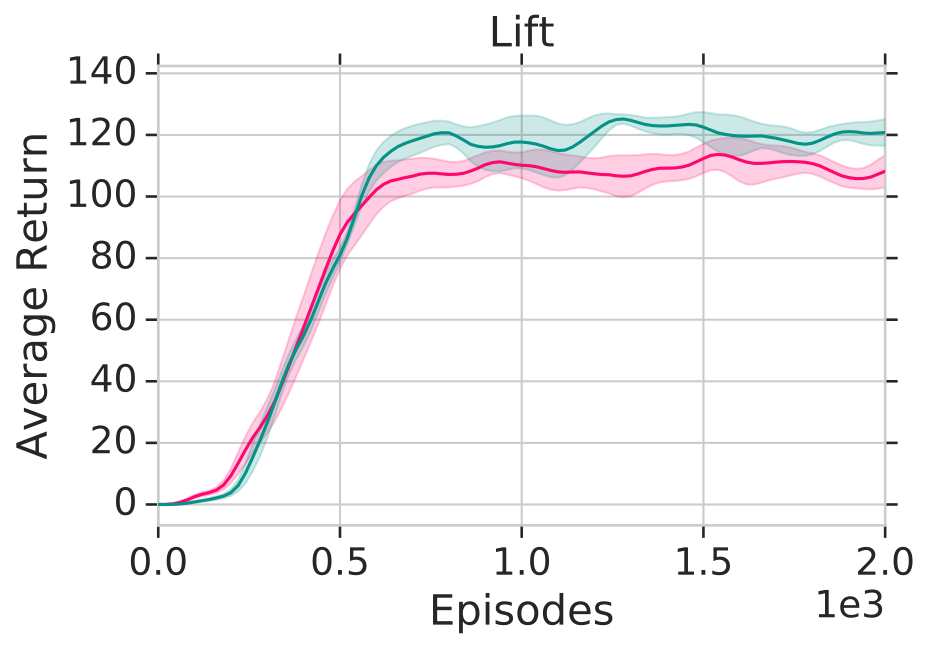}
        \includegraphics[width = .235\textwidth]{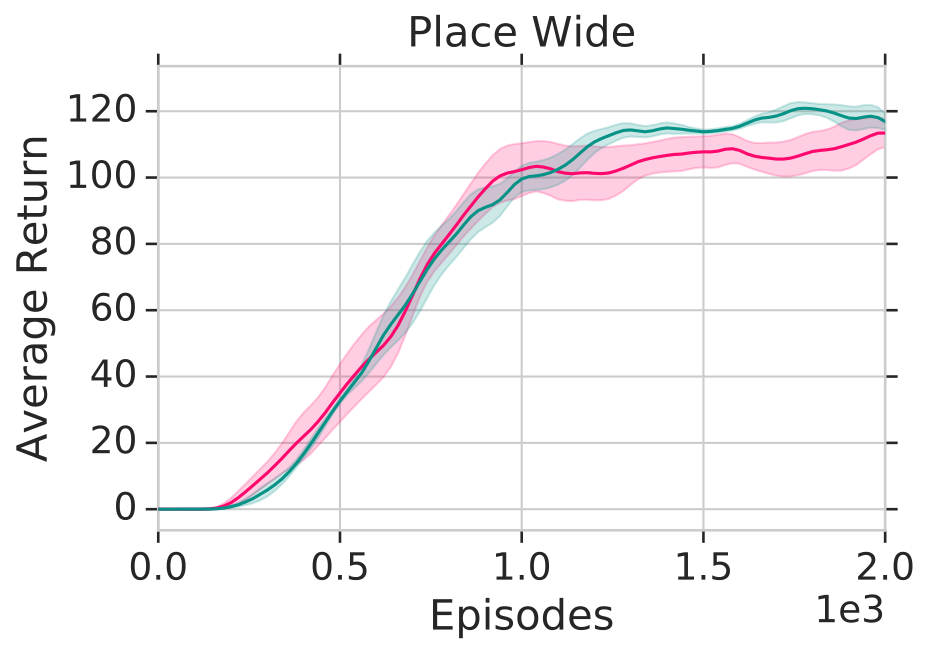}\\
        \includegraphics[width = .235\textwidth]{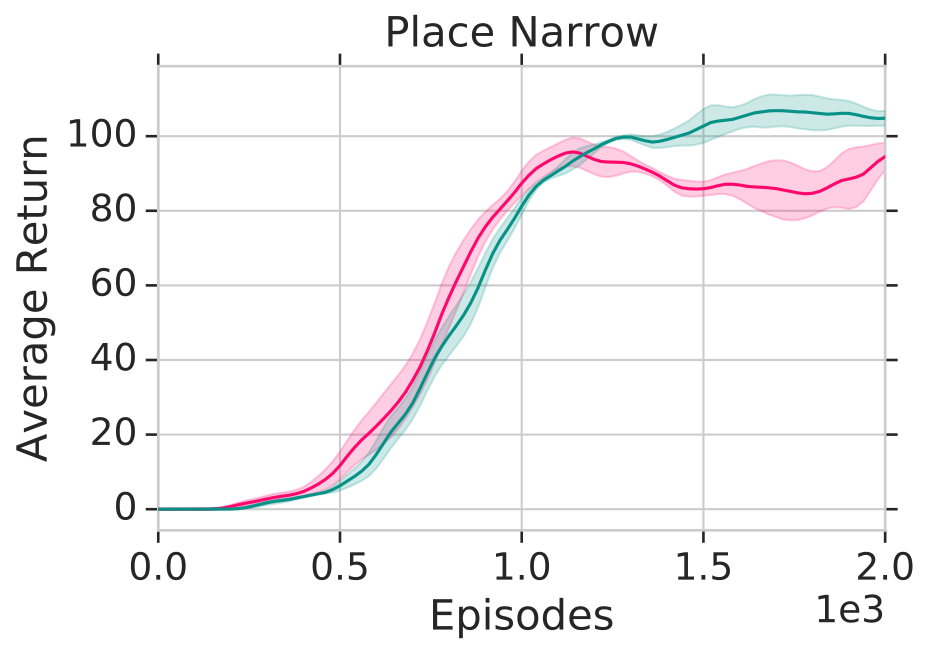}
        \includegraphics[width = .235\textwidth]{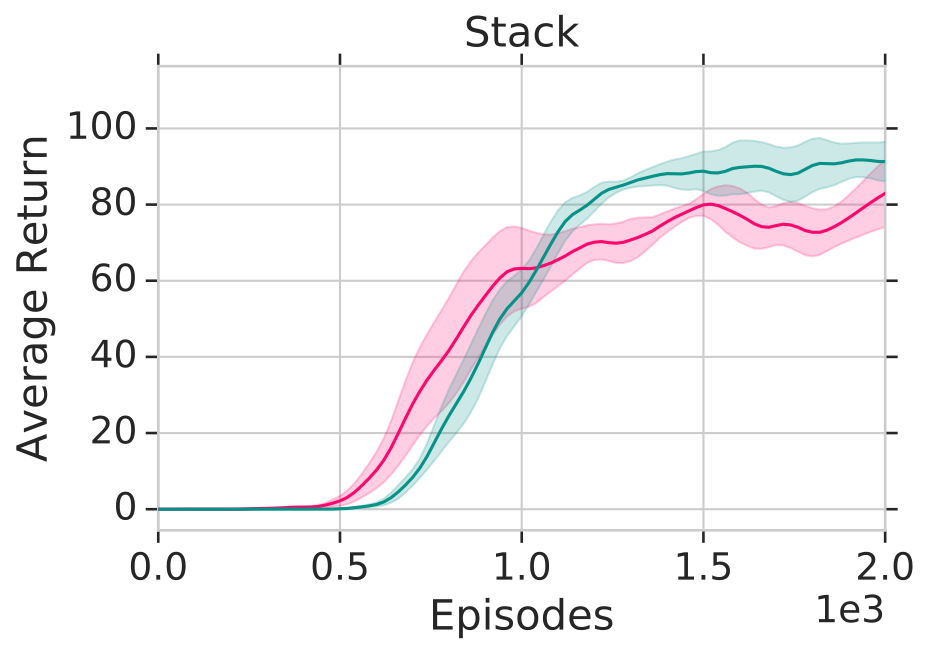}
        \includegraphics[width = .235\textwidth]{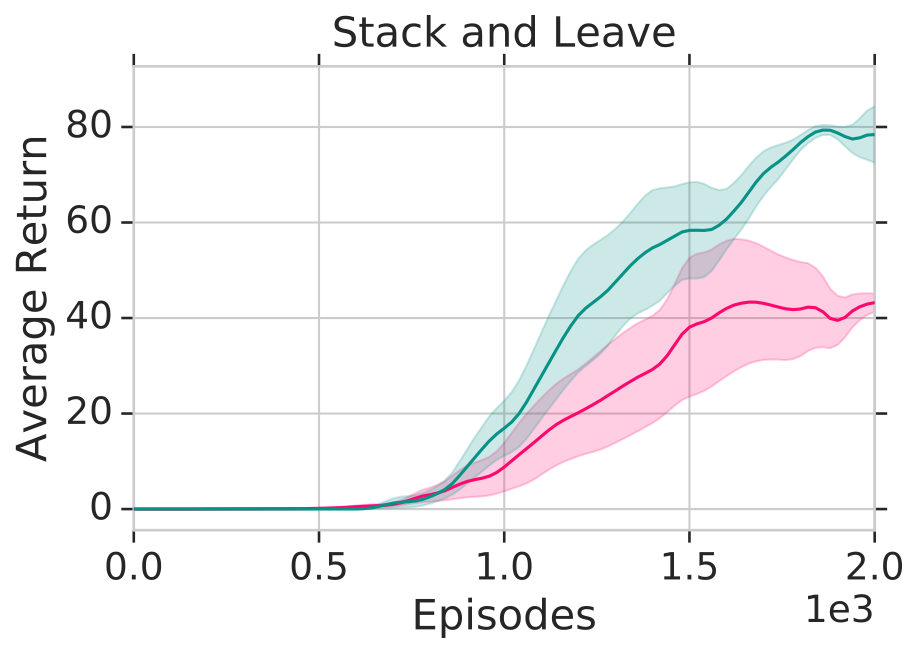}
	\end{tabular}
    \caption{Complete results on multi-task block stacking with and without conditioning termination conditions on tasks.}
	\label{fig:terminations}
\end{figure*}

The complete results for all experiments with task-agnostic terminations can be found in Figure \ref{fig:terminations}.
    
\subsection{Off-Policy Option Learning}
\label{sec:off_policy_exps}
In order to train in a more on-policy regime, we reduce the size of the replay buffer by two orders of magnitude and increase the ratio between data generation (actor steps) and data fitting (learner steps) by one order of magnitude. The resulting algorithm is run without any additional hyperparameter tuning to provide an insight into the effect of conditioning on action probabilities under options in the inference procedure. 
We can see that in the on-policy case the impact of this change is less pronounced. Across all cases, we were unable to generate significant performance gains by including action conditioning into the inference procedure.

\begin{figure*}[h]
	\centering
	\begin{tabular}{c}
        \includegraphics[width = 0.98\textwidth]{results/png/gym_ac.png}\\
        \includegraphics[width = 0.98\textwidth]{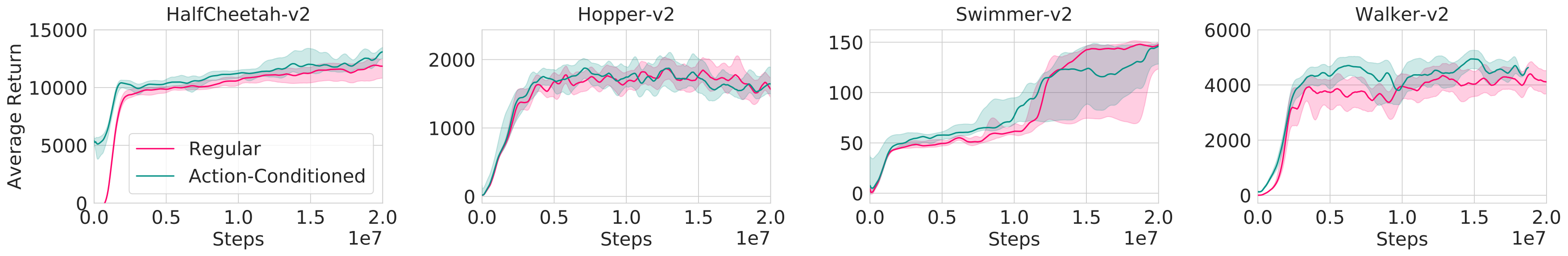}\\
	\end{tabular}
    \caption{Complete results on OpenAI gym with and without conditioning component probabilities on past executed actions. For the off-policy (top) and on-policy case (bottom). The on-policy approaches uses data considerably less efficiently and the x-axis is correspondingly adapted. }
	\label{fig:ac_onpol}
\end{figure*}

The complete results for all experiments with and without the action-conditional inference procedure can be found in Figure \ref{fig:ac_onpol}.

\subsection{Trust-region Constraints}
\label{sec:trust_region_exps}
\begin{figure*}[h]
	\centering
	\begin{tabular}{cccc}
        \includegraphics[width = .235\textwidth]{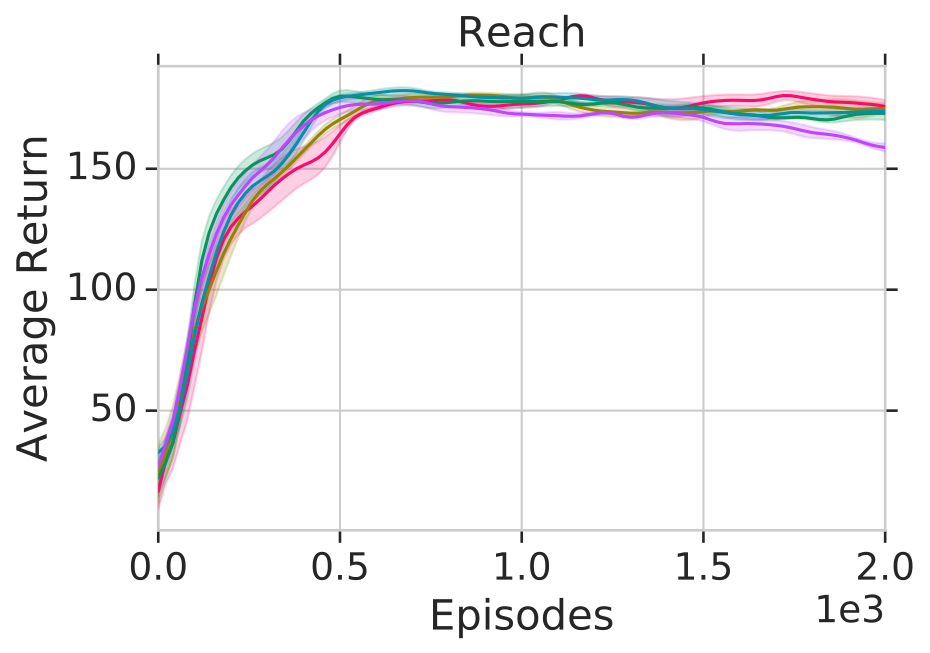} 
        \includegraphics[width = .235\textwidth]{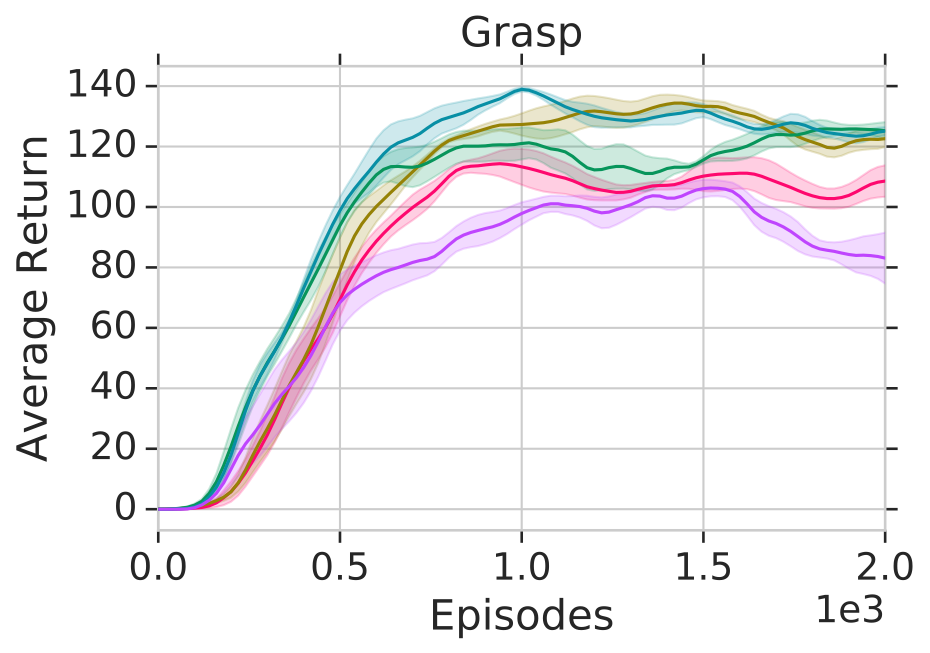}
        \includegraphics[width = .235\textwidth]{results/png/trustregion/pile1_tr_2.png}
        \includegraphics[width = .235\textwidth]{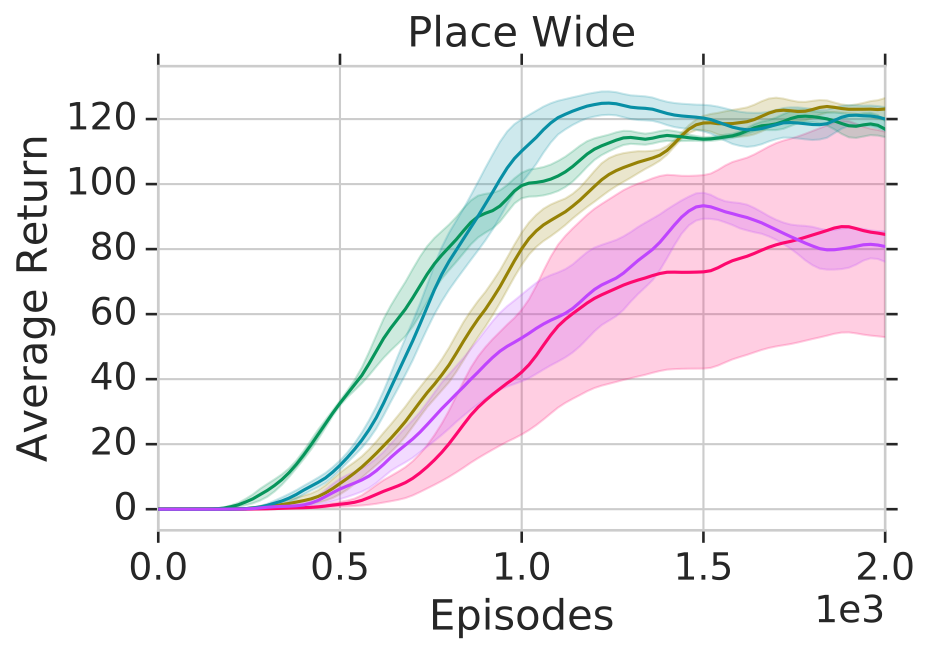}\\
        \includegraphics[width = .235\textwidth]{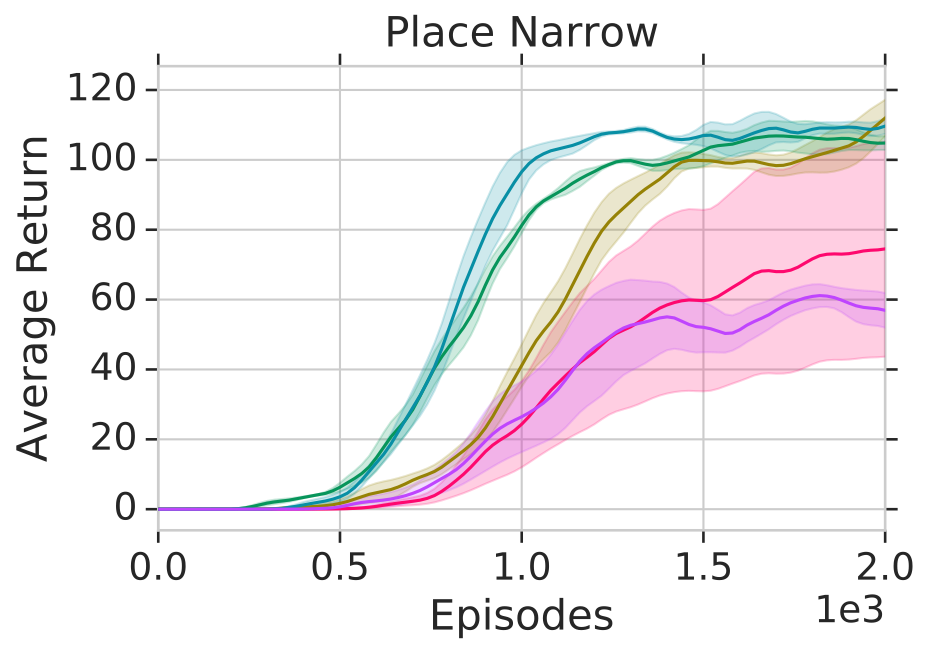}
        \includegraphics[width = .235\textwidth]{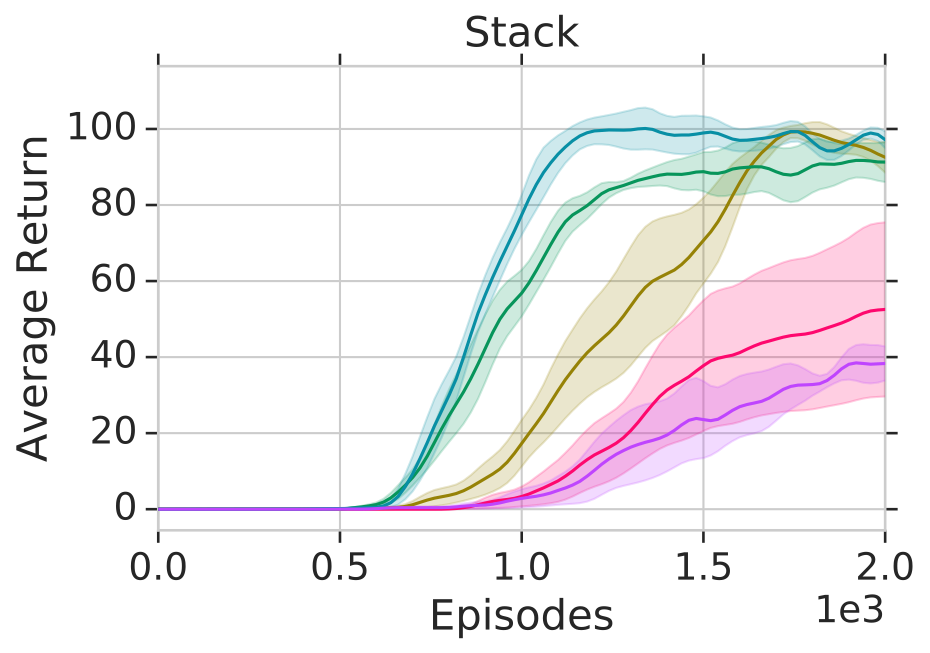}
        \includegraphics[width = .235\textwidth]{results/png/trustregion/pile1_tr_6.png}
	\end{tabular}
    \caption{Complete results on block stacking with varying trust-region constraints for both termination conditions $\beta$ and the high-level controller $\pi_C$. }
	\label{fig:trustregions_all}
\end{figure*}

The complete results for all trust-region ablation experiments can be found in Figure \ref{fig:trustregions_all}.

With the exception of very high or very low constraints, the approach trains robustly, but performance drops considerably when we remove the constraint fully.

\subsection{Single Time-Step vs Multi Time-Step Inference}
\label{sec:sampling_exps}

\begin{figure*}[h]
	\centering
	\begin{tabular}{c}
        \includegraphics[width = .98\textwidth]{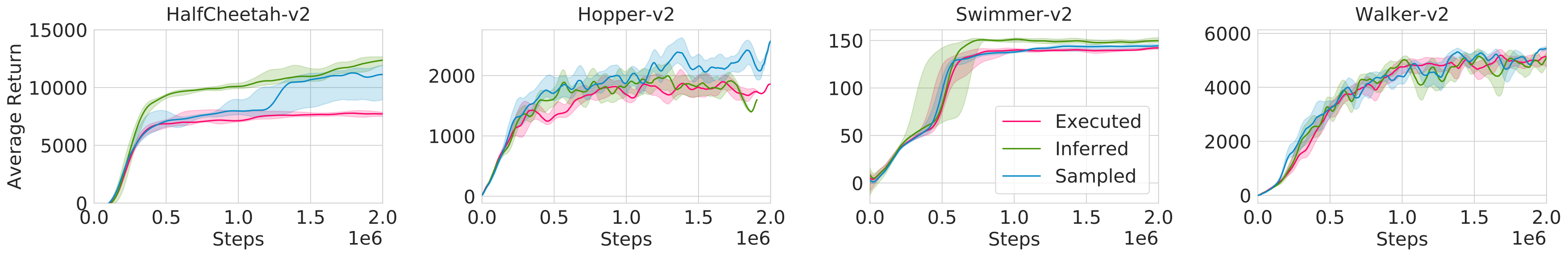} 
	\end{tabular}
    \caption{Ablation results comparing inferred options with sampled options during learning (sampled) and during execution (executed). The ablation is run with five actors instead of a single one as used in the OpenAI gym experiments in order to generate results faster. }
	\label{fig:sampling_options}
\end{figure*}

To investigate the impact of probabilistic inference of posterior option distributions $\pi_H(o_t|h_t)$ along the whole sampled trajectory instead of using sampling-based approximations until the current timestep, we perform additional ablations displayed in Figure \ref{fig:sampling_options}. Note that we are required to perform probabilistic inference for at least one step to use backpropagation through the inference step to update our policy components. Any completely sampling-based approach would require a different policy optimizer (e.g. via likelihood ratio or reparametrization trick) which would introduce additional compounding effects.

We compare \met~with an ablated version where we do not compute the option probabilities along the trajectory following Equation \ref{eq:dynamic1} but instead use an approximation with only concrete option samples propagating across timesteps for all steps until the current step. 
To generate action samples, we therefore sample options for every timestep along a trajectory without keeping a complete distribution over options and sample actions only from the active option at every timestep. 
To determine the likelihood of actions and options for every timestep, we rely on Equation \ref{eq:optiontransitions} based the sampled options of the previous timestep.
By using samples and the critic-weighted update procedure from Equation \ref{eq:objective_pi}, we can only generate gradients for the policy for the current timestep instead of backpropagating through the whole inference procedure. We find that using both samples from executed options reloaded from the buffer as well as new samples during learning can reduce performance depending on the domain. However, in the Hopper-v2 environment, sampling during learning performs slightly better than inferring options.

\section{Additional Experiment Details \label{app:details}}

\subsection{OpenAI Gym Experiments}

All experiments are run with asynchronous learner and actors. We use a single actor and report performance over the number of transitions generated. Following \citep{wulfmeier2019regularized}, both \met~and RHPO use different biases for the initial mean of all options or mixture components - distributed between minimum and maximum action output. This provides a small but non-negligible benefit and supports specialization of individual options. In line with our baselines (DAC \citep{zhang2019dac}, IOPG \citep{smith2018inference}, Option Critic \citep{bacon2017option}) we use 4 options or mixture components for the OpenAI gym experiments. We run all experiments with 5 samples and report variance and mean. All experiments are run with a single actor in a distributed setting. The variant with limited switches limits to 2 switches over a sequence length of 8. Lower and higher values led to comparable results.


\begin{table}[h!]
\begin{center}
 \begin{tabular}{c||c|c|c} 
 Hyperparameters & \met & RHPO & MPO \\
 \hline
 Policy net & \multicolumn{3}{c}{256-256}\\ 
 Number of actions samples& \multicolumn{3}{c}{20}\\
 Q function net & \multicolumn{3}{c}{256-256}\\
 Number of components & \multicolumn{2}{c}{4} & NA\\
 $\epsilon$ &\multicolumn{3}{c}{0.1} \\
 $\epsilon_{\mu}$ & \multicolumn{3}{c}{5e-4} \\
 $\epsilon_{\Sigma}$ & \multicolumn{3}{c}{5e-5}\\
 $\epsilon_{\alpha}$ & \multicolumn{2}{c}{1e-4} & NA\\
 $\epsilon_{t}$ & 1e-4 & \multicolumn{2}{c}{NA}\\
 Discount factor ($\gamma$) & \multicolumn{3}{c}{0.99} \\
 Adam learning rate & \multicolumn{3}{c}{3e-4} \\
 Replay buffer size & \multicolumn{3}{c}{2e6} \\
 Target network update period & \multicolumn{3}{c}{200}\\
 Batch size & \multicolumn{3}{c}{256}\\
 Activation function & \multicolumn{3}{c}{elu}\\
 Layer norm on first layer & \multicolumn{3}{c}{Yes}\\
 Tanh on output of layer norm & \multicolumn{3}{c}{Yes}\\
 Tanh on actions (Q-function) & \multicolumn{3}{c}{Yes} \\
 Sequence length & \multicolumn{3}{c}{8}   
\end{tabular}
\end{center}
\caption{Hyperparameters - OpenAI gym}
\label{tab:Single}
\end{table}

\subsection{Action and Temporal Abstraction Experiments}

Shared across all algorithms, we use 3-layer convolutional policy and Q-function torsos with [128, 64, 64] feature channels, [(4, 4), (3, 3), (3, 3)] as kernels and stride 2.
For all multitask domains, we build on information asymmetry and only provide task information as input to the high-level controller and termination conditions to create additional incentive for the options to specialize. The Q-function has access to all observations (see the corresponding tables in this section).
We follow \citep{riedmiller2018learning,wulfmeier2019regularized} and assign rewards for all possible tasks to trajectories when adding data to the replay buffer independent of the generating policy. 

\begin{table}[h!]
\begin{center}
 \begin{tabular}{c||c|c|c} 
 Hyperparameters & \met & RHPO & MPO \\
 \hline
 Policy torso \\(shared across tasks) & \multicolumn{2}{c}{256} & 512\\ 
 \begin{tabular}{@{}c@{}} Policy task-dependent \\ heads \end{tabular} &\multicolumn{2}{c}{ \begin{tabular}{@{}c@{}} 100  (cat.) \end{tabular}} & 200 \\ 
 \begin{tabular}{@{}c@{}}  Policy shared \\ heads \end{tabular} & \multicolumn{2}{c}{\begin{tabular}{@{}c@{}} 100  (comp.) \end{tabular}} & NA \\
 \begin{tabular}{@{}c@{}} Policy task-dependent \\ terminations \end{tabular} &\begin{tabular}{@{}c@{}} 100 \\ (term.) \end{tabular} & NA & NA \\ 
 $\epsilon_{\mu}$ & \multicolumn{3}{c}{1e-3} \\
 $\epsilon_{\Sigma}$ & \multicolumn{3}{c}{1e-5}\\
 $\epsilon_{\alpha}$ & \multicolumn{2}{c}{1e-4} & NA\\
 $\epsilon_{t}$ & 1e-4 & \multicolumn{2}{c}{NA}\\
 Number of action samples& \multicolumn{3}{c}{20}\\
 Q function torso \\(shared across tasks) & \multicolumn{3}{c}{400}\\
 Q function head \\(per task)& \multicolumn{3}{c}{300}\\
 Number of components & \multicolumn{2}{c}{\begin{tabular}{@{}c@{}} number of tasks \end{tabular}} & NA\\
 Replay buffer size & \multicolumn{3}{c}{1e6} \\
 Target network \\ update period & \multicolumn{3}{c}{500}\\
 Batch size & \multicolumn{3}{c}{256}\\
\end{tabular}
\end{center}
\caption{Hyperparameters. Values are taken from the OpenAI gym experiments with the above mentioned changes.}
\label{tab:MPOMulti}
\end{table}

\paragraph{Stacking}

The setup consists of a Sawyer robot arm mounted on a table and equipped with a Robotiq 2F-85 parallel gripper. In front of the robot there is a basket of size 20x20 cm which contains three cubes with an edge length of 5 cm (see Figure \ref{fig:evs}).

The agent is provided with proprioception information for the arm (joint positions, velocities and torques), and the tool center point position computed via forward kinematics. For the gripper, it receives the motor position and velocity, as well as a binary grasp flag. It also receives a wrist sensor's force and torque readings. Finally, it is provided with three RGB camera images at $64 \times 64$ resolution. At each timestep, a history of two previous observations (except for the images) is provided to the agent, along with the last two joint control commands. The observation space is detailed in Table \ref{tab:sawyer_observations}. All stacking experiments are run with 50 actors in parallel and reported over the current episodes generated by any actor. Episode lengths are up to 600 steps.

The robot arm is controlled in Cartesian velocity mode at 20Hz. The action space for the agent is 5-dimensional, as detailed in Table \ref{tab:pile_actions}. The gripper movement is also restricted to a cubic volume above the basket using virtual walls.

\begin{table}[h!]
\caption{Action space for the Sawyer Stacking experiments.}
\label{sawyer-action-table2}
\vskip 0.15in
\begin{center}
\begin{small}
\begin{tabular}{lccc}
\toprule
Entry & Dims & Unit & Range \\
\midrule
Translational Velocity in x, y, z & 3 & m/s & [-0.07, 0.07] \\
Wrist Rotation Velocity & 1 & rad/s & [-1, 1] \\
Finger speed & 1 & tics/s & [-255, 255] \\
\bottomrule
\end{tabular}
\end{small}
\end{center}
\vskip -0.1in
\label{tab:pile_actions}
\end{table}

\begin{table}[h!]
\caption{Observations for the Sawyer Stacking experiments. The TCP's pose is represented as its world coordinate position and quaternion. In the table, $m$ denotes meters, $rad$ denotes radians, and $q$ refers to a quaternion in arbitrary units ($au$).}
\label{tab:sawyer_observations}
\vskip 0.15in
\begin{center}
\begin{small}
\begin{tabular}{lccc}
\toprule
Entry & Dims & Unit & History \\
\midrule
Joint Position (Arm) & 7 & rad & 2 \\
Joint Velocity (Arm) & 7 & rad/s & 2 \\
Joint Torque (Arm) & 7 & Nm & 2 \\
Joint Position (Hand) & 1 & tics & 2 \\
Joint Velocity (Hand) & 1 & tics/s & 2 \\
Force-Torque (Wrist) & 6 & N, Nm & 2 \\
Binary Grasp Sensor & 1 & au & 2 \\
TCP Pose & 7 & m, au & 2 \\
Camera images & $3 \times 64 \times $ & R/G/B value & 0 \\
 & $64 \times 3$ & & \\
Last Control Command & 8 & rad/s, tics/s & 2 \\
\bottomrule
\end{tabular}
\end{small}
\end{center}
\vskip -0.1in
\end{table}

\begin{equation}
stol(v, \epsilon, r) =
\begin{cases}
  1 &\text{iff} \ |v| < \epsilon \\
  1 - tanh^2( \frac{atanh(\sqrt{0.95})}{r} |v|) &\text{else}
\end{cases}
\label{eq:shaped_tolerance}
\end{equation}

\begin{equation}
slin(v, \epsilon_{min}, \epsilon_{max}) =
\begin{cases}
  0 &\text{iff} \ v < \epsilon_{min} \\
  1 &\text{iff} \ v > \epsilon_{max} \\
  \frac{v - \epsilon_{min}}{\epsilon_{max} - \epsilon_{min}}  &\text{else}
\end{cases}
\label{eq:shaped_tolerance2}
\end{equation}

\begin{equation}
btol(v, \epsilon) =
\begin{cases}
  1 &\text{iff} \ |v| < \epsilon  \\
  0 &\text{else}
\end{cases}
\end{equation}

\begin{itemize}
    \item \textit{REACH(G)}: $stol(d(TCP, G), 0.02, 0.15)$: \\
    Minimize the distance of the TCP to the green cube.
    \item \textit{GRASP}: \\
    Activate grasp sensor of gripper ("inward grasp signal" of Robotiq gripper)
    \item \textit{LIFT(G)}: $slin(G, 0.03, 0.10)$ \\
    Increase z coordinate of an object more than 3cm relative to the table.
    \item \textit{PLACE\_WIDE(G, Y)}: $stol(d(G, Y + [0,0,0.05]), 0.01, 0.20)$\\
    Bring green cube to a position 5cm above the yellow cube.
    \item \textit{PLACE\_NARROW(G, Y)}: $stol(d(G, Y + [0,0,0.05]), 0.00, 0.01)$: \\
    Like PLACE\_WIDE(G, Y) but more precise.
    \item \textit{STACK(G, Y)}: $btol(d_{xy}(G, Y), 0.03) * btol(d_z(G, Y) + 0.05, 0.01) * (1 - \textit{GRASP})$ \\
    Sparse binary reward for bringing the green cube on top of the yellow one (with 3cm tolerance horizontally and 1cm vertically) and disengaging the grasp sensor.
    \item \textit{STACK\_AND\_LEAVE(G, Y)}: $ stol(d_z(TCP, G)+0.10, 0.03, 0.10) * \textit{STACK(G, Y)}$ \\
    Like STACK(G, Y), but needs to move the arm 10cm above the green cube.
\end{itemize}

\paragraph{Ball-In-Cup}

This task consists of a Sawyer robot arm mounted on a pedestal. A partially see-through cup structure with a radius of 11cm and height of 17cm is attached to the wrist flange. Between cup and wrist there is a ball bearing, to which a yellow ball of 4.9cm diameter is attached via a string of 46.5cm length (see Figure \ref{fig:evs}).

Most of the settings for the experiment align with the stacking task. The agent is provided with proprioception information for the arm (joint positions, velocities and torques), and the tool center point and cup positions computed via forward kinematics. It is also provided with two RGB camera images at $64 \times 64$ resolution. At each timestep, a history of two previous observations (except for the images) is provided to the agent, along with the last two joint control commands. The observation space is detailed in Table \ref{tab:sawyer_bic_observations}. All BIC experiments are run with 20 actors in parallel and reported over the current episodes generated by any actor. Episode lengths are up to 600 steps.

The position of the ball in the cup's coordinate frame is available for reward computation, but not exposed to the agent.
The robot arm is controlled in joint velocity mode at 20Hz. The action space for the agent is 4-dimensional, with only 4 out of 7 joints being actuated, in order to avoid self-collision. Details are provided in Table \ref{tab:pile_actions}.

\begin{table}[h!]
\caption{Action space for the Sawyer Ball-in-Cup experiments.}
\label{sawyer-action-table3}
\vskip 0.15in
\begin{center}
\begin{small}
\begin{tabular}{lccc}
\toprule
Entry & Dims & Unit & Range \\
\midrule
Rotational Joint Velocity \\
for joints 1, 2, 6 and 7 & 4 & rad/s & [-2, 2] \\
\bottomrule
\end{tabular}
\end{small}
\end{center}
\vskip -0.1in
\label{tab:bic_actions}
\end{table}

\begin{table}[h!]
\caption{Observations for the Sawyer Ball-in-Cup experiments. In the table, $m$ denotes meters, $rad$ denotes radians, and $q$ refers to a quaternion in arbitrary units ($au$). Note: the joint velocity and command represent the robot's internal state; the 3 degrees of freedom that were fixed provide a constant input of 0.}
\label{tab:sawyer_bic_observations}
\vskip 0.15in
\begin{center}
\begin{small}
\begin{tabular}{lcc}
\toprule
Entry & Dims & Unit \\
\midrule
Joint Position (Arm) & 7 & rad \\
Joint Velocity (Arm) & 7 & rad/s \\
TCP Pose & 7 & m, au \\
Camera images & $2 \times 64 \times 64 \times 3$ & R/G/B value \\
Last Control Command & 7 & rad/s \\
\bottomrule
\end{tabular}
\end{small}
\end{center}
\vskip -0.1in
\end{table}

Let $B_A$ be the Cartesian position in meters of the ball in the cup's coordinate frame (with an origin at the center of the cup's bottom), along axes $A \in \{x, y, z\}$.

\begin{itemize}
    \item \textit{CATCH}: $0.17 > B_z > 0$ and $||B_{xy}||_2 < 0.11$ \\
    Binary reward if the ball is inside the volume of the cup.
    \item \textit{BALL\_ABOVE\_BASE}: $B_z > 0$ \\
    Binary reward if the ball is above the bottom plane of the cup.
    \item \textit{BALL\_ABOVE\_RIM}: $B_z > 0.17$\\
    Binary reward if the ball is above the top plane of the cup.
    \item \textit{BALL\_NEAR\_MAX}: $B_z > 0.3$\\
    Binary reward if the ball is near the maximum possible height above the cup.
    \item \textit{BALL\_NEAR\_RIM}: $1 - tanh^2( \frac{atanh(\sqrt{0.95})}{0.5} \times ||B_{xyz}-(0,0,0.17)||_2)$\\
    Shaped distance of the ball to the center of the cup opening (0.95 loss at a distance of 0.5).
\end{itemize}

\subsection{Pre-training and Sequential Transfer Experiments} 
The sequential transfer experiments are performed with the same settings as their multitask equivalents. However, they rely on a pre-training step in which we take all but the final task in each domain and train \met~to pre-train options which we then transfer with a new high-level controller on the final task. Fine-tuning of the options is enabled as we find that it produces slightly better performance. Only data used for the final training step is reported but all both approaches were trained for the same amount of data during pretraining until convergence. 
The variant with limited switches limits to 4 switches over a sequence length of 16. 
\subsection{Locomotion experiments}
\label{sec:locomotion_details}

\begin{figure}[h]
	\centering
        \includegraphics[width = .3\textwidth, clip, trim = 0 0 0 2mm]{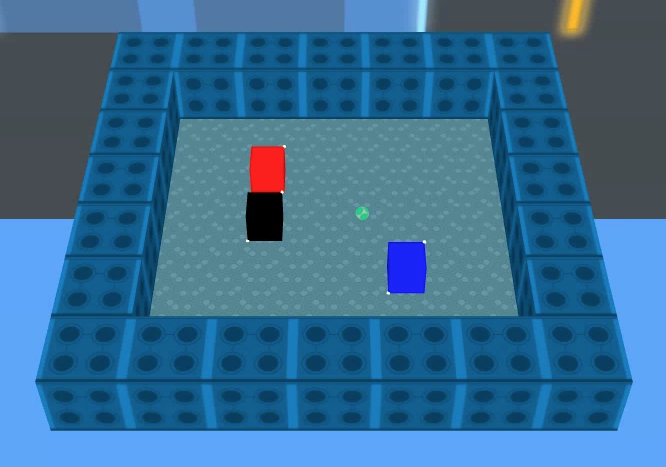}
        \includegraphics[width = .3\textwidth, clip, trim = 0 0 0 0]{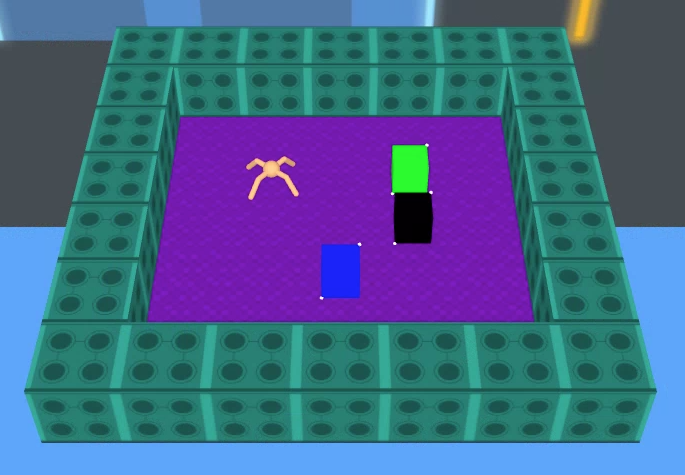}
        \includegraphics[width = .3\textwidth, clip, trim = 0 0 0 3mm]{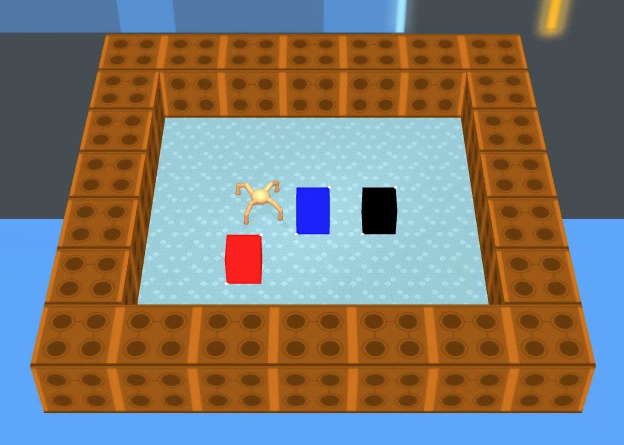}
    \caption{The environment used for simple locomotion tasks with Ball (top), Ant (center) and Quadruped (bottom).}
	\label{fig:locomotion_env}
\end{figure}

Figure~\ref{fig:locomotion_env} shows examples of the environment for the different bodies used.
In addition to proprioceptive agent state information (which includes the body height, position of the end-effectors, the positions and velocities of its joints and sensor readings from an accelerometer, gyroscope and velocimeter attached to its torso), the state space also includes the ego-centric coordinates of all target locations and a categorical index specifying the task of interest. Table \ref{tab:predicate_observations} contains an overview of the observations and action dimensions for this task.
The agent receives a sparse reward of $+60$ if part of its body reaches a square surrounding the predicate location, and $0$ otherwise.
Both the agent spawn location and target locations are randomized at the start of each episode, ensuring that the agent must use both the task index and target locations to solve the task.

\begin{table}[h!]
\caption{Observations for the \textit{go to one of 3 targets} task with Ball, Ant, and Quadruped.}
\label{tab:predicate_observations}
\vskip 0.15in
\begin{center}
\begin{small}
\begin{tabular}{lccc}
\toprule
Entry & Dimensionality  \\
\midrule
Task Index & 3 \\
Target locations & 9 \\
Proprioception (Ball) & 16 \\
Proprioception (Ant) & 41 \\
Proprioception (Quad) & 57 \\
Action Dim (Ball) & 2 \\
Action Dim (Ant) & 8 \\
Action Dim (Quad) & 12 \\
\bottomrule
\end{tabular}
\end{small}
\end{center}
\vskip -0.1in
\end{table}

\section{Additional Derivations \label{app:additiona_derivations}}

In this section we explain the derivations for training option policies with options parameterized as Gaussian distributions. Each policy improvement step is split into two parts: non-parametric and parametric update.

\subsection{Non-parametric Option Policy Update \label{app:nonparam}}
In order to obtain the non-parametric policy improvement we optimize the following equation:

\begin{equation*}
  \begin{aligned}
  & \max_q \mathbb{E}_{h_t \sim p(h_t)} \big[ \mathbb{E}_{a_t,o_t \sim q} \big[Q_\phi(s_t,a_t,o_t)] \big] \\  
  & s.t. \mathbb{E}_{h_t \sim p(h_t)} \big[ \textrm{KL}(q(\cdot|h_t) , \pi_\theta(\cdot|h_t) ) \big] < \epsilon_E\\
  & s.t. \mathbb{E}_{h_t \sim p(h_t)} \big[ \mathbb{E}_{q(a_t,o_t | h_t)} \big [1 \big] \big] = 1.
  \end{aligned}
\end{equation*}

for each step $t$ of a trajectory, where $h_{t} = \{s_t, a_{t-1}, s_{t-1},... a_0,s_0\}$ represents the history of states and actions and $p(h_t)$ describes the distribution over histories for timestep $t$, which in practice are approximated via the use of a replay buffer $\mathcal{D}$. When sampling $h_t$, the state $s_t$ is the first element of the history. The inequality constraint describes the maximum allowed KL divergence between intermediate update and previous parametric policy, while the equality constraint simply ensures that the intermediate update represents a normalized distribution.

Subsequently, in order to render the following derivations more intuitive, we replace the expectations and explicitly use integrals. 
The Lagrangian $L(q,\eta,\gamma)$ can now be formulated as
\begin{align}\label{eq:completelagrange1}
L(q,\eta,\gamma) = \iiint p(h_t) q(a_t,o_t|h_t) Q_\phi(s_t,a_t,o_t) \\ 
\hfill\nonumber \diff o_t \diff a_t \diff h_t  \\
\nonumber  +\eta \bigg(\epsilon_E - \iiint p(h_t)q(a_t,o_t|h_t)
\log \frac{q(a_t,o_t|h_t)}{\pi_\theta(a_t,o_t|h_t)}\\
\nonumber \hfill \diff o_t \diff a_t \diff h_t \bigg)\\
\nonumber + \gamma\left(1 - \iiint p(h_t) q(a_t,o_t|h_t)\diff o_t \diff a_t \diff h_t \right).
\end{align}

Next to maximize the Lagrangian with respect to the primal variable $q$, we determine its derivative as,
\begin{align*}
\frac{\partial L(q,\eta,\gamma)}{{\partial q}} = Q_\phi(a_t,o_t,s_t) - \eta\log q(a_t,o_t|h_t) \\
\nonumber+\eta\log \pi_\theta(a_t,o_t|h_t) - \eta - \gamma.
\end{align*}
In the next step, we can set the left hand side to zero and rearrange terms to obtain
\begin{align*}
q(a_t,o_t|h_t) = \pi_\theta(a_t,o_t|h_t)\exp\left(\frac{Q_\phi(s_t,a_t,o_t)}{\eta}\right)\\
\exp\left(-\frac{\eta+\gamma}{\eta}\right).
\end{align*}
The last exponential term represents a normalization constant for $q$, which we can formulate as
\begin{eqnarray}
\begin{aligned}
\frac{\eta+\gamma}{\eta}= \log\bigg(&\iint \pi_\theta(a_t,o_t|h_t)\label{eq:norm}\\
&\exp\left(\frac{Q_\phi(s_t,a_t,o_t)}{\eta}\right)\diff o_t \diff a_t\bigg).
\end{aligned}
\end{eqnarray}
In order to obtain the dual function $g(\eta)$, we insert the solution for the primal variable into the Lagrangian in Equation \ref{eq:completelagrange1} which yields 
\begin{align*}
L(q,\eta,\gamma) = \iiint p(h_t)q(a_t,o_t|h_t) Q_\phi(s_t,a_t,o_t) \\
 \hfill \diff o_t \diff a_t \diff h_t \\
+\eta\bigg(\epsilon_E - \iiint p(h_t) q(a_t,o_t|h_t)\\
\log {\scriptscriptstyle \frac{\pi_\theta(a_t,o_t|h_t)\exp\left(\frac{Q_\phi(s_t,a_t,o_t)}{\eta}\right)\exp\left(-\frac{\eta+\gamma}{\eta}\right)}{\pi_\theta(a_t,o_t|h_t)} } \diff o_t \diff a_t \diff h_t \bigg) \\
+ \gamma\left(1 - \iiint p(h_t) q(a_t,o_t|h_t)\diff o_t \diff a_t \diff h_t \right).
\end{align*}

We expand the equation and rearrange to obtain
\begin{align*}
L(q,\eta,\gamma) &=\iiint p(h_t) q(a_t,o_t|h_t) Q_\phi(s_t,a_t,o_t) \\
& \hfill ~~~~~~~~~~~~~~~~~~~~~~~~~~~~~~~~~~~~~~~~~~~~~~~~~~~~~ \diff o_t \diff a_t \diff h_t  \\
& - \eta\iiint p(h_t) q(a_t,o_t|h_t)\Big[\frac{Q_\phi(s_t,a_t,o_t)}{\eta} \\
&\hfill+ \log \pi_\theta(a_t,o_t|h_t) - \frac{\eta+\gamma}{\eta}\Big]\diff o_t \diff a_t \diff h_t  \\& + \eta\epsilon_E
 +\eta\iiint p(h_t) q(a_t,o_t|h_t)\\
&~~~~~~~~~~~~~~~~~~~~~~~~~~~~~~~~~~\log \pi_\theta(a_t,o_t|h_t)\diff o_t \diff a_t \diff h_t  \\ & + \gamma\left(1 - \iiint p(h_t) q(a_t,o_t|h_t)\diff o_t \diff a_t \diff h_t \right).
\end{align*}

In the next step, most of the terms cancel out and after additional rearranging of the terms we obtain
\begin{align*}
L(q,\eta,\gamma) = \eta\epsilon_E + \eta\int p(h_t)\frac{\eta+\gamma}{\eta}\diff h_t.
\end{align*}
We have already calculated the term inside the integral in Equation \ref{eq:norm}, which we now insert to obtain
\begin{align}
    g(\eta) =& \min_q L(q,\eta,\gamma)\label{eq:dual_eta}\\
    =& \eta\epsilon_E+\eta\int p(h_t)\log\bigg(\iint \pi_\theta(a_t,o_t|h_t)\nonumber\\
    &~~~~~~~~~~~~~~~~~~~\exp\left(\frac{Q_\phi(s_t,a_t,o_t)}{\eta}\right)\diff o_t \diff a_t \bigg)\diff h_t \nonumber\\
    =& \eta\epsilon_E+\eta \mathbb{E}_{h_t\sim p(h_t)} \Big[ \log\bigg(\mathbb{E}_{a_t,o_t \sim \pi_\theta}\Big[\nonumber\\
    &~~~~~~~~~~~~~~~~~~~~ \exp\left(\frac{Q_\phi(s_t,a_t,o_t)}{\eta}\right)\Big]\bigg)\Big]. \nonumber
\end{align}
The dual in Equation \ref{eq:dual_eta} can finally be minimized with respect to $\eta$ based on samples from the replay buffer and policy.

\subsection{Parametric Option Policy Update \label{app:param}}

After obtaining the non-parametric policy improvement, we can align the parametric option policy to the current non-parametric policy.
As the non-parametric policy is represented by a set of samples from the parametric policy with additional weighting, this step effectively employs a type of critic-weighted maximum likelihood estimation. In addition, we introduce regularization based on a distance function $\mathcal{T}$ which has a trust-region effect for the update and stabilizes learning.

\begin{eqnarray*}
\begin{aligned}
\theta_{new} =&\arg \min_{\theta} \mathbb{E}_{h_t \sim p(h_t)}\Big[ \mathrm{KL}\big( q(a_t,o_t | h_t) \| \pi_{\theta}(a_t,o_t  | h_t) \big) \Big] \\
=&\arg \min_{\theta} \mathbb{E}_{h_t \sim p(h_t)}\Big[ \mathbb{E}_{a_t,o_t \sim q} \Big[ \log q(a_t,o_t  | h_t) \\
\nonumber &~~~~~~~~~~~~~~~~~~~~~~~~~~~~~~~~~~~~~~~~~~~~~~~-\log \pi_{\theta}(a_t,o_t  | h_t)  \Big] \Big] \\
=& \argmax_{\theta}
\mathbb{E}_{h_t \sim p(h_t), a_t, o_t  \sim q}\Big[ \log \pi_{\theta}(a_t, o_t  |h_t) \Big], \quad \\&\textrm{s.t.}\:\mathbb{E}_{h_t \sim p(h_t)} \Big[ \mathcal{T}(\pi_{\theta_{new}}(\cdot|h_t) | \pi_{\theta}(\cdot|h_t)) \Big] < \epsilon_M,
\end{aligned}
\end{eqnarray*}

where $h_t \sim p(h_t)$ is a trajectory segment, which in practice sampled from the dataset $\mathcal{D}$, $\mathcal{T}$ is an arbitrary distance function between the new policy and the previous policy. $\epsilon_M$ denotes the allowed change for the policy. We again employ Lagrangian Relaxation to enable gradient based optimization of the objective, yielding the following primal:

\begin{align}\label{eq:lagrange}
\max_\theta \min_{\alpha > 0}  L(\theta,\alpha) = \mathbb{E}_{ h_t \sim p(h_t), a_t,o_t \sim q}\Big[ \log \pi_{\theta}(a_t,o_t|h_t) \Big] \nonumber\\
+\alpha\Big(\epsilon_M - \mathbb{E}_{h_t \sim p(h_t)} \big[ \mathcal{T}(\pi_{\theta_{new}}(\cdot|h_t) , \pi_{\theta}(\cdot|h_t))\big]\Big).
\end{align}

We can solve for $\theta$ by iterating the inner and outer optimization programs independently. In practice we find that it is most efficient to update both in parallel.

We also define the following distance function between old and new option policies
\begin{align*}
\mathcal{T}(\pi_{\theta_{new}}(\cdot|h_t),\pi_{\theta}(\cdot|h_t)) = \mathcal{T}_{H}(h_t) +\mathcal{T}_{T}(h_t) + \mathcal{T}_{L}(h_t)
\end{align*}

\begin{align*}
\mathcal{T}_{H}(h_t) = \textrm{KL}(\textrm{Cat}(\{\alpha_{\theta_{new}}^{j}(h_t)\}_{j=1...M}) \|\\
\textrm{Cat}(\{\alpha_{\theta}^{j}(h_t)\}_{j=1...M}))\\
\mathcal{T}_{T}(h_t) = \frac{1}{M}\sum_{j=1}^{M} \textrm{KL}(\textrm{Cat}(\{\beta_{\theta_{new}}^{ij}(h_t)\}_{j=1...2}) \| \\ \textrm{Cat}(\{\beta_{\theta}^{ij}(h_t)\}_{j=1...2}))\\
\mathcal{T}_{L}(h_t) = \frac{1}{M}\sum_{j=1}^{M}\textrm{KL}(\mathcal{N}(\mu^j_{\theta_{new}}(h_t),\Sigma^j_{\theta_{new}}(h_t)) \| \\ \mathcal{N}(\mu^j_{\theta}(h_t),\Sigma^j_{\theta}(h_t)))
\end{align*}

where $\mathcal{T}_{H}$ evaluates the KL between the categorical distributions of the high-level controller, $\mathcal{T}_{T}$ is the average KL between the categorical distributions of the all termination conditions, and $\mathcal{T}_{L}$ corresponds to the average KL across Gaussian components. In practice, we can exert additional control over the convergence of model components by applying different $\epsilon_M$ to different model parts (high-level controller, termination conditions, options).

\subsection{Transition Probabilities for Option and Switch Indices}
The transitions for option $o$ and switch index $n$ are given by:
\begin{align}
    p(o_t,n_t|s_t,o_{t-1},n_{t-1}) = ~~~~~~~~~~~~~~~~~~~~~~~~~~~~~~~~~~~~~~ &\nonumber\\ 
    \begin{cases} \label{eq:optionswitch_transitions}
      (1-\beta(s_t,o_{t-1})) & \text{if } n_{t} = n_{t-1}, o_t=o_{t-1} \\
      \beta(s_t,o_{t-1}) \pi^C(o_t|s_t)      & \text{if } n_{t} = n_{t-1}+1\\
      0                                      & \text{otherwise}
    \end{cases}
\end{align}



\end{document}